 \documentclass[final,3p,times,twocolumn]{elsarticle}

\usepackage{amssymb}
\usepackage{hyperref}
\usepackage{graphicx}
\usepackage{subcaption}
\usepackage{lscape}
\usepackage{amsmath}
\usepackage{multirow}
\usepackage[figuresright]{rotating}
\usepackage{lineno}
\usepackage{algorithm} 
\usepackage{algpseudocode} 
\usepackage{comment}
\usepackage{lineno}

\begin{document}

\begin{frontmatter}



\title{Performance Evaluation of Deep Transfer Learning on Multiclass Identification of Common Weed Species in Cotton Production Systems}


\author[label1]{Dong Chen}
\author[label2]{Yuzhen Lu}
\author[label1]{Zhaojiang Li}
\author[label3]{Sierra Young}

\address{Yuzhen Lu (yzlu@abe.msstate.edu) is the corresponding author}
\address[label1]{Department of Mechanical Engineering, Michigan State University, East Lansing, MI 48824, USA}
\address[label2]{Department of Agricultural and Biological Engineering, Mississippi State University, Mississippi State 39762, MS, USA}
\address[label3]{Department of Biological and Agricultural Engineering, North Carolina State University, Raleigh, NC 27695, USA}

\begin{abstract}
Precision weed management offers a promising solution for sustainable cropping systems through the use of chemical-reduced/non-chemical robotic weeding techniques, which apply suitable control tactics to individual weeds. Therefore, accurate identification of weed species plays a crucial role in such systems to enable precise, individualized weed treatment. Despite recent progress, the development of a robust weed identification and localization system in the presence of unstructured field conditions remains a serious challenge, requiring supervised modeling using large volumes of annotated data. This paper makes a first comprehensive evaluation of deep transfer learning (DTL) for identifying common weeds specific to cotton (\textit{Gossypium hirsutum} L.) production systems in southern United States (U.S.). A new dataset for weed identification was created, consisting of 5187 color images of 15 weed classes collected under natural lighting conditions and at varied weed growth stages, in cotton fields (primarily in Mississippi and North Carolina) during the 2020 and 2021 field seasons. We evaluated 27 state-of-the-art deep learning models through transfer learning and established an extensive benchmark for the considered weed identification task. DTL achieved high classification accuracy of F1 scores exceeding 95\%, requiring reasonably short training time (less than 2.5 hours) across models. ResNet101 achieved the best F1-score of 99.1\% whereas 14 out of the 27 models achieved F1 scores exceeding 98.0\%. However, the performance on minority weed classes with few training samples was less satisfactory for models trained with a conventional, unweighted cross entropy (CE) loss function. To address this issue, a weighted cross entropy (WCE) loss function was adopted, which achieved substantially improved accuracies for minority weed classes (e.g., the F1-scores for MnasNet and EfficientNet-b2 on the Spurred Anoda weed increased from 20\% to 80\% and 40\% to 80\%, respectively). Furthermore, a deep learning-based cosine similarity metrics was employed to analyze the similarity among weed classes, assisting in the interpretation of classifications. Both the codes (\url{https://github.com/Derekabc/CottonWeeds}) for model benchmarking and the weed dataset (\url{https://www.kaggle.com/yuzhenlu/cottonweedid15}) of this study are made publicly available, which expect to be be a valuable resource for future research in weed identification and beyond.

\end{abstract}


\end{frontmatter}


\section{Introduction}
\label{sec:intro}

Weeds are critical threats to crop production; potential crop yield loss due to weeds is estimated at 43\% on a global scale \cite{oerke2006crop}. In cotton production, poor weed management can lead to yield loss of up to 90\% \cite{manalil2017weed}. Weeds control is traditionally performed through machines or by hand weeding. With the advent of trans-genetic, glyphosate-tolerant crops since 1996, over 90\% of the U.S. farm lands for field crops such as cotton, are planted with herbicide-resistant seeds \cite{usda2015genetically}. The weed control has become predominantly reliant on herbicide application \cite {duke2015perspectives, pandey2021autonomy}. Intensive, blanket, broadcast application of herbicides, however, has adverse environmental impacts and facilitates the evolution of herbicide-resistant weeds (e.g., Palmer Amaranth and Waterhemp), which in turn substantially increases management costs \cite{norsworthy2012reducing}.

Precision weed management (PWM) has recently emerged as a promising solution for sustainable, effective weed control, which incorporates sensors, computer systems and robotics into cropping systems \cite{young2014future}. By recognizing the biological attributes of different weed species, PWM enables precise and minimum necessary treatments according to site-specific demand and targeting individual weeds or a small cluster \cite{gerhards2003real}, which can lead to significant reduction in the consumption of herbicides and other resources. For instance, a robotic weeder can spray a particular type or volume of herbicide or use mechanical weeder or lasers to treat specific weed species, avoiding unnecessary application to crops, bare soil or plant residuals \cite{barnes2021opportunities}. Therefore, successful implementation of integrated, precise weed control strategies relies on accurate identification, localization and monitoring of weeds. Currently, machine vision and robotic technology for automated weed control have been demonstrated in certain speciality crops \cite{fennimore2019robotic}. However, commercial-scale applicability to row crops such as cotton in varying growing conditions has yet to be evaluated or demonstrated. Lack of a robust machine vision system capable of weed recognition with accuracy exceeding 95\% in unstructured field conditions has been identified as one of the most critical technological bottlenecks towards full realization of automated weeding \cite{westwood2018weed}. The key to addressing this bottleneck thus lies in the development of image analysis and modeling algorithms of high and robust performance.

Image analysis methods based on the extraction of color and texture features, followed by thresholding or supervised modeling, are widely used for weed classification and detection \cite{wang2019review, meyer2008verification}. A variety of color indices that accentuate plant greenness have been proposed for separating weeds from soil backgrounds \cite{meyer2008verification, woebbecke1995color}. The color indices that are developed from empirical observations are however not robust enough in dealing with images acquired under variable field lighting conditions \cite{hamuda2016survey}. In \cite{bawden2017robot}, texture features including local binary pattern and covariance features were used to perform weed classification, and the extracted features that were applied to a robotic platform, achieved an accuracy of 92.3\% on the dataset containing 40 images of 6 weed species. Local shape and edge orientation features were used in \cite{ahmad2018visual} for discriminating monocot and dicot weeds, which achieved an overall accuracy of 98.4\% based on AdaBoost with Naïve Bayes. In  \cite{bakhshipour2018evaluation}, Fourier descriptors and invariant moments were extracted and fed into support vector machine for classifying four common weeds in sugarbeet fields, resulting in an accuracy of 93.3\% accuracy. Despite promising results, the aforementioned color or texture feature-based approaches require engineering hand-crafted features for given weed detection/classification tasks, which may not adapt satisfactorily to a more diverse set of imaging conditions.

Recently, data-driven methods such as deep learning (DL), e.g., convolutional neural networks (CNNs), have been researched for weed classification and detection \cite{hasan2021survey}. CNNs are able to capture spatial and temporal dependencies of images through the use of shared-weight filters and can be trained end-to-end without explicit feature extraction \cite{o2015introduction}, empowering neural networks to adaptively discover the underlying class-specific patterns and the most discriminative features. In \cite{dyrmann2016plant}, a CNN model, which was trained on a dataset containing 10413 images with 22 plant species at early growth stage, achieved a classification accuracy of up to 98\%.  A graph-based DL architecture with multi-scale graph representations was developed in  \cite{hu2020graph} for weed classification, achieving an accuracy of 98.1\%  on the DeepWeeds dataset \cite{olsen2019deepweeds}. While successful, training such DL models from scratch are very time-consuming and resource-intensive, requiring high-performance computation units and large-scale, high-quality annotated image datasets, which may not be readily available. 

Transfer learning, a methodology that aims at transferring the knowledge across domains, can greatly reduce the training time and the dependence on massive training data by reusing already trained models for new problems \cite{zhuang2020comprehensive}. Therefore, deep transfer learning (DTL, i.e., transferring DL models) only involves fine-tuning model parameters using new datasets in the target domain. DTL has recently been investigated for weed identification. In \cite{olsen2019deepweeds}, two pretrained DL models were fined-turned and tested on the DeepWeeds dataset, achieving average accuracies above 95\%. In \cite{espejo2020improving}, the authors found that fune-tuning DL models on agricultural datasets helped reduce training epochs while improving model accuracies. They fine-tuned four DL models on the Plant Seedlings Dataset \cite{Giselsson2017} and the Early Crop Weeds Dataset \cite{espejo2020towards} and improved the classification accuracy by 0.51\% to 1.89\%, respectively.  
In \cite{suh2018transfer}, six pretrained DL models wer adopted to classify sugarbeet and volunteer potato images and achieved the best accuracy of up to 98.7\%. In \cite{ahmad2021performance}, three pre-trained CNN models, were used for weed classification, achieving 98.8\% accuracy in classifying four weed species in corn and soybean fields. These studies, however, only experimented a small number of DL models. Given active developments in DL model architectures \cite{khan2020survey}, it would be beneficial to the research community to evaluate a broad range of state-of-the-art DL models on weed identification, so as to facilitate informed selection of high-performance models in terms of accuracy, training time, model complexity, and inference speed. 

Despite transfer learning strategies, having large volumes of annotated image data is highly desirable for powering DL models in visual categorization tasks \cite{sun2017revisiting}. Currently, the dearth of such datasets remains a crucial hurdle for exploiting the potential of DL and advancing machine vision systems for precision agriculture \cite{lu2020survey, eden}. In weed detection, to achieve high accuracy and robustness requires a dataset that provides adequate representation of important weed species and accounts for the variations associated with environmental factors (e.g., soil types and characteristics, field light, shadows) as well as growth-stage-related morphological or physiological variations. Recently, Lu and Young (2020) reviewed 15 publicly available weed  image datasets dedicated to weed control \cite{lu2020survey}, such as DeepWeeds \cite{olsen2019deepweeds}, Early crop weed dataset \cite{espejo2020towards}, Open Plant Phenotyping Database \cite {leminen2020open}, among some others. Most of these datasets target a small number of weed species, with images acquired from a single growth season in geographically similar field sites. No image datasets of weeds specific to cotton production systems have been published so far. 

In this paper, we present a new weed dataset collected in cotton fields in multiple southern U.S. states over the two consecutive seasons of 2020 and 2021. We establish a comprehensive benchmark of a large set of DL architectures for weed classification on the new dataset. This research is expected to provide a valuable reference for future research on developing machine vision systems for cotton weed control and beyond. The contributions of this paper are highlighted as follows:

\begin{enumerate}
\item The presentation of a unique, diverse weed dataset\footnote{\url{https://www.kaggle.com/yuzhenlu/cottonweedid15}} consisting of 5187 images of 15 weed classes specific to the U.S. cotton production systems.
\item A comprehensive evaluation and benchmark of 27 state-of-the-art DL models\footnote{\url{https://github.com/Derekabc/CottonWeeds}} through transfer learning for multi-class weed identification. 
\item A novel DL-based cosine similarity metric for assisting in the interpretation of DL output and a weighted loss function for improving classification accuracies for minority weed classes.
\end{enumerate}


\section{Materials and Methods}
\label{sec:data_collection}

\subsection{Cotton Weed Dataset}
RGB (Red-Green-Blue) images of weed plants were collected from cotton fields using either smartphones or hand-held digital color cameras. For the sake of image diversity, following the recommendations in \cite{lu2020survey}, images were captured from different view angles, under natural field light conditions, at varying stages of weed growth, at different locations across the U.S. cotton belt states (primarily in North Carolina and Mississippi). Regular visits to cotton fields were conducted throughout June to August in the growing seasons of 2020 and 2021 for weed image collection. In 2020, images were mainly acquired in the cotton fields of North Carolina State University research stations, including Central Crops Research Station (Clayton, NC), Upper Coastal Plain Research Station (Rocky Mount, NC) and Cherry Research Farm (Goldsboro, NC). In 2021, more weed images were acquired in cotton fields of R. R. Foil Plant Science Research Center (Starkville, MS) and Black Belt Experiment Station (Brooksville, MS) of Mississippi State University. To create a diverse, large-scale dataset, weed scientists at different institutions were invited to participate in the image collection effort. A Google form\footnote{\url{https://forms.gle/zr9wa1uu7qHTFiK2A}} was created and shared for uploading weed images and associated metadata (e.g., weed species, field sites, weather conditions).

\begin{figure*}[!ht]
\centering
\includegraphics[width=0.85\textwidth]{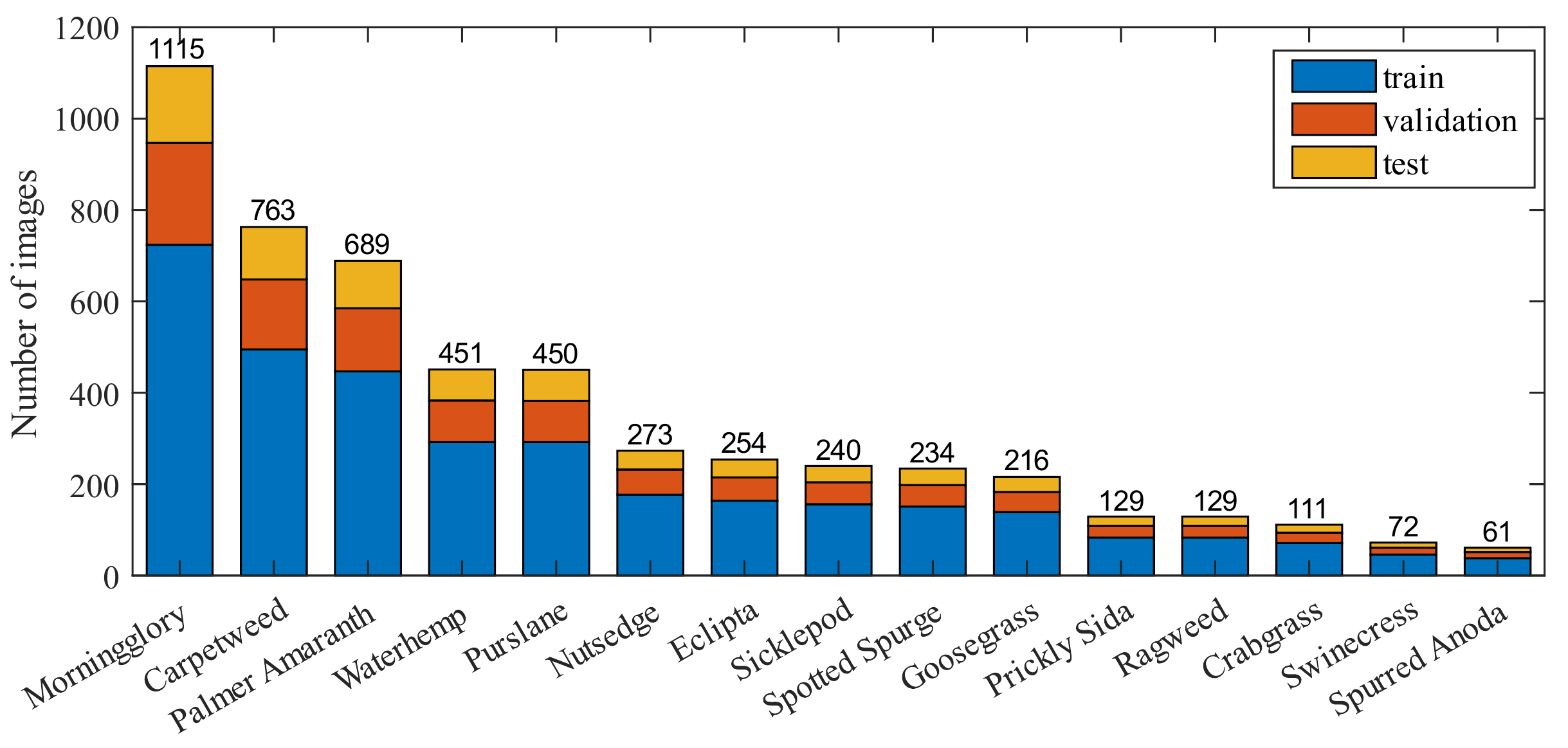}
\caption{Bar plot of the cotton weed dataset. Images per weed class are randomly partitioned into 65\%-20\%-15\% splits of train, validation and test subsets, respectively. Numbers above the bars represent the total number of images for the corresponding weed classes.}
\label{fig:bar_plot}
\end{figure*}

The acquired images were first annotated for weed species by weed experts during image submission through the Google from, and the received images were then annotated by trained individuals, and the final annotations were examined again by experts to ensure annotation quality. The images with multiple classes of weeds were cropped so that each resultant image contained a single class of weeds. The weed classes were defined by common names of weed plants. At the time of writing, the entire dataset contains more than 10000 images for over 50 weed species, which will be documented in detail in a future study. Here the weed dataset used for benchmarking DL models consists of a total of 5187 images for 15 common weed classes. The image number for each weed class is shown in Fig.~\ref{fig:bar_plot}. It should be noted that all weed classes, except Morningglory, correspond to single weed species. The images of different Morningglory species (e.g., Ivy Morninglory, Pitted Morningglory, Entireleaf Morningglory and Tall Morningglory) were grouped together as a single weed class, because of their similarity in weed management. Overall the weed classes including Morningglorg, Carpetweed, Palmer Amaranth, Waterhemp and Purlane are the four major classes in terms of image number, as opposed to weed species like Crabgrass, Swinecress and Spurred Anoda, corresponding to minority classes. 
It is clear that the present dataset has unbalanced classes. The class imbalance generally poses a challenge to machine learning modeling, which will be discussed in Sections~\ref{subsec:wce} and \ref{subsec:wce_r}.

\begin{figure*}[!ht]
  \centering
  \includegraphics[width=0.84\textwidth]{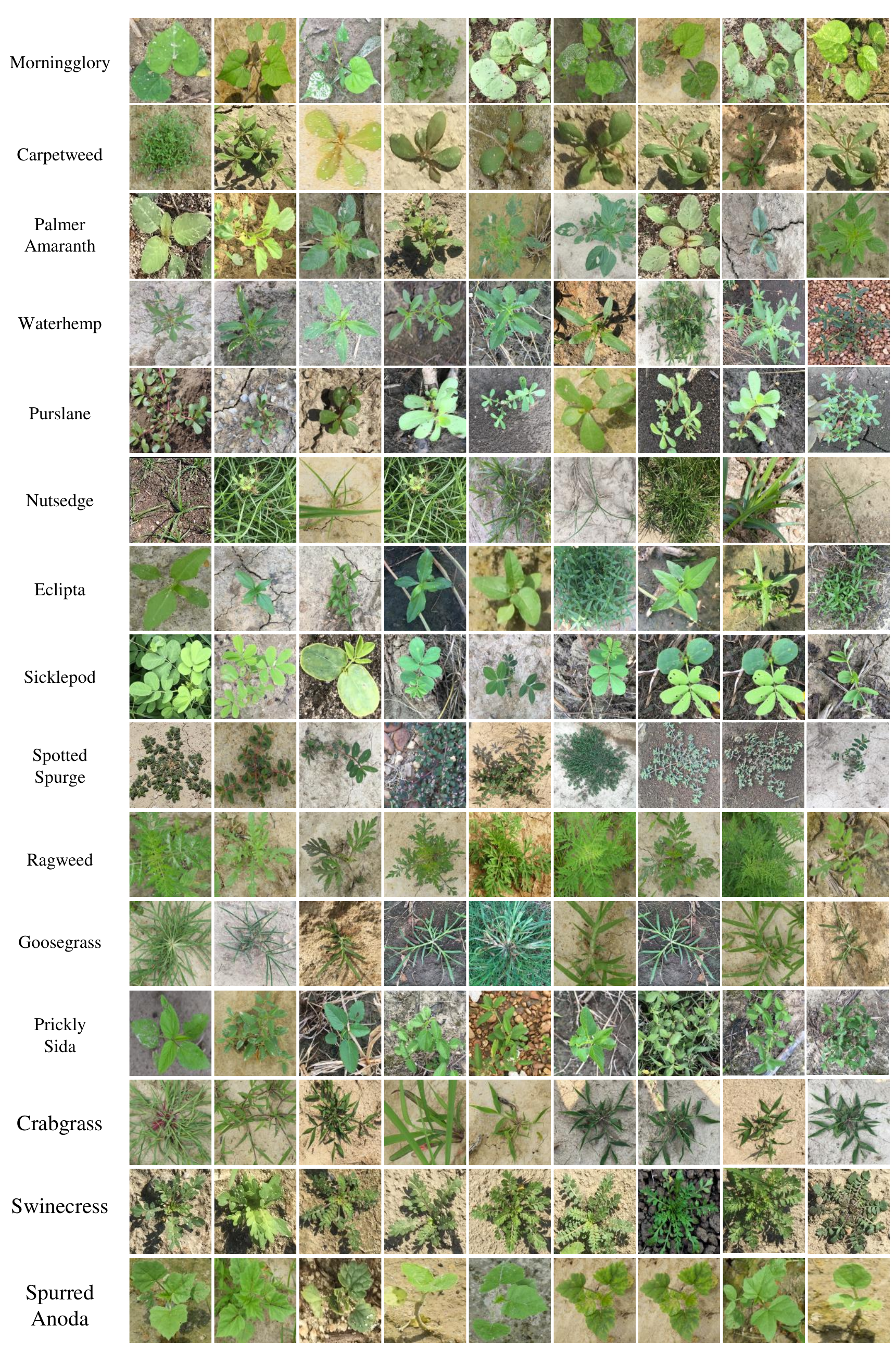}
  \caption{Example images from the cotton weed dataset. Each row displays randomly selected images from each of the 15 weed classes.}
  \label{fig:weeds}
\end{figure*}

Fig.~\ref{fig:weeds} shows example images from the cotton weed dataset. The images within the same weed class have large variations in leaf color and morphology, soil background and field light conditions, which are desirable for building models robust to image conditions or dataset shift. The image variations vary among weed classes; despite distinct identifying characteristics, some weed classes exhibit relatively high similarities in the plant morphology. For instance, some young Morningglory and Spurred Anoda seedlings have similar, broad leaves, and the latter is also similar to Prickly Sida in terms of toothed leaf margins. Goosegrass and Crabgrass are both grassy weeds that grow prostrate on the ground, with similar leaf shapes. Palmer Amaranth and Waterhemp that are both pigweed species may look similar and are difficult to distinguish from each other. These similarities may contribute to errors in weed identification by DL models. A quantitative DL-based similarity measure along with a similarity matrix will be discussed in later sections (see Sections~\ref{subsec:smi} and \ref{subsec:smir}) to characterize the similarity among weed classes.

\subsection{Transfer learning}
\label{subsec:tl}
Deep transfer learning (DTL) starts with a pre-trained DL model on a large-scale dataset (e.g., ImageNet \cite{deng2009imagenet}) and then fine-tunes the model on a new dataset from the specific domain of interest \cite{zhuang2020comprehensive}. For the weed classification in this study, we replace the last fully-connected (FC) layers of DL models with a layer that has 15 neurons corresponding to the same number of weed classes in the cotton weed dataset. 

Literature review was conducted to select appropriate DL models for weed identification. The main selection criterion was the demonstrated performance of models in visual categorization tasks in the computer vision community and the availability of their source-code implementations. As a result, a suite of 27 state-of-the-art CNN models of different architectures, as summarized in Table~\ref{tab:models} , were selected for classifying the cotton weed images here. Some of them including Xcpetion \cite{chollet2017xception}, VGG16 \cite{simonyan2014very}, ResNet50 \cite{he2016deep}, InceptionV3 \cite{szegedy2016rethinking} and DenseNet \cite{huang2017densely}  has recently been evaluated for classifying weeds in other cropping systems \cite{espejo2020towards, olsen2019deepweeds, ahmad2021performance}. The majority of these models, such as EfficientNet \cite{tan2019efficientnet} and MnasNet \cite{tan2019mnasnet}, remain to be evaluated for weed classification tasks.

The DL models were trained with a conventional cross entropy (CE) loss function as follows:
\begin{linenomath*}
\begin{equation} \label{eqn:ce_loss}
CE_{loss} = - \sum_{i=1}^C t_i \log(p_i),
\end{equation}
\end{linenomath*}
where $p_i\in\mathbb{R}^{15}$ is a vector of a Softmax output layer \cite{goodfellow2018softmax} indicating the probability of the 15 weed classes. Here $C=15$ is the number of weed classes, and $t_i$ denotes the true probability and is defined as:
\begin{linenomath*}
\begin{equation}
    t_i = 
    \begin{cases}
          1, &\text{if } i \text{ is the true label}; \\
          0, &  \text{otherwise}.
    \end{cases}
\end{equation}
\end{linenomath*}

For model development and evaluation, the cotton weed dataset was randomly partitioned into three subsets: 65\% for training, 20\% for validation and 15\% for testing, as shown in Fig.~\ref{fig:bar_plot}. All training and validation images were resized to $512 \times 512$ pixels in size before being fed into DL models (images of $256 \times 256$ and $1080 \times 1080$ pixels were also examined, but the size of $512 \times 512$ pixels was found to be better in terms of accuracy and speed). The image pixel intensities per color channel were normalized to the range of $[-1, 1]$ for enhanced image recognition performance \cite{koo2017image}. In addition, for better model accuracy, real-time data augmentation was conducted by randomly rotating images in the range of $[-180^o, +180^o]$ and random flipping during the training process.

Because of the random nature of dataset partition, it would be desirable to run model training and testing multiple times for obtaining a reliable estimate of model performance \cite{raschka2018model}. In this study, DL models were trained with 5 replications, with different random seeds that were shared by all the models, and the mean accuracies on test data were computed for performance evaluation. All the models were trained for 50 epochs (that were found sufficient for modeling the weed data) with the SGD (stochastic gradient descent) optimizer and a momentum of 0.9. The learning rate was initially set to be 0.001, and dynamically decreased by a factor of 0.1 every 7 epochs for stabilizing model training. The DL framework Pytorch (version 1.9) with Torchvision (version 0.10.0) \cite{paszke2019pytorch} were used for model training, in which a multiprocessing package\footnote{\url{https://docs.python.org/3/library/multiprocessing.html}} was employed with 32 CPU cores to speed up the training. The experiments were performed on an Ubuntu 20.04 server with an AMD 3990X 64-Core CPU and a GeForce RTX 3090Ti GPU (24 GB GDDR6X memory). Readers are referred to the open-source codes\footnote{Code at: \url{https://github.com/Derekabc/CottonWeeds}}, for detailed implementation of transfer learning for the 27 DL models.

\subsection{Performance Metrics}
The performance of the DL models in weed identification was evaluated in terms of number of model parameters, training and inference times, confusion matrix and F1-score.

\subsubsection{Number of Model Parameters}
In this study, pretrained DL models were fined tuned by updating all the model parameters for the weed classification task. Thus the number of model parameters refer to all the weights (and biases) in the network that are updated/learnt during the training process through back-propagation. 
The parameter number is a direct measure of model complexity: networks with a larger number of parameters potentially require greater deployment memory and incur longer training and inference times (see Subsection 2.3.2).

\subsubsection{Training and Inference Times}
The training time is the time required to train a DL model with prescribed model configurations and computing resources. The training time depends on factors such as model architecture, number of model parameters, data size, hyper-parameters, DL framework as well as computing hardware. The training time is an important consideration where development time and resources are constrained. 

A trained DL model is to be used to make predictions (also known as \textit{inference}). The inference time (i.e., latency) is one crucial aspect in deploying DL models for real-time applications (e.g., in-field weed identification). It is the time that a trained DL model takes to make a prediction given an image input. In this paper, for reliable estimation, the inference time was measured as the average time needed to predict 30 weed images randomly selected from the testing dataset.

\subsubsection{Confusion Matrix and F1-score}
The confusion matrix on testing images, which provides the accuracy for each class while revealing detailed misclassifications, was presented to show the classification for individual weed classes. The classification accuracy was measured in terms of F1 score. For the multi-class weed classification, the micro-averaged F1 score \cite{yang1999re} was calculated as the classification accuracy. In micro-averaging, the per-class classifications are aggregated across classes to compute the micro-averaged precision \textit{P} and recall \textit{R} by counting the total true positives, false negatives and false positives, and then a harmonic combination of \textit{P} and \textit{R}, i.e., Micro-F1, as follows:
\begin{linenomath*}
\begin{equation} \label{eqn:f1}
F1  = \frac{2 PR}{P + R}.
\end{equation}
\end{linenomath*}

\subsection{Weighted Cross Entropy Loss}
\label{subsec:wce}
The CE loss function defined in Eqn.~\ref{eqn:ce_loss} does not account for the class imbalance encountered in the cotton weed dataset (Fig.~\ref{fig:weeds}) in this study. Training with the CE loss may result in large classification errors for minority weed classes (e.g., Spurred Annoda). To mitigate this issue, a weighted cross entropy (WCE) \cite{phan2020resolving} loss function was introduced, performing re-weighting according to image numbers for each weed class as follows:
\begin{linenomath*}
\begin{equation} \label{eqn:wce_loss}
WCE_{loss} = - \sum_{i=1}^C  w_i \text{ } t_i \log(p_i),
\end{equation}
\end{linenomath*}
where $w$ is a weighting vector that assigns individualized penalty to each class, preferentially placing larger weights on minority classes. The  conventional CE loss without considering class imbalance corresponds to a weighting vector of ones (the CE column in Table~\ref{tab:weights}). In this study, an inverse-proportion weighting strategy \cite{phan2020resolving} was adopted to assign the weight to the $i$th weed class as follows:
\begin{linenomath*}
\begin{equation}\label{eqn:weights}
    w_i = \frac{N_{max}}{N_i},  
\end{equation}
\end{linenomath*}
where $N_i$ denotes the number of images for the $i$th weed class and $N_{max}$ represents the maximum number of images among classes, i.e., 1115 (for Morningglory). As a result, weed classes with fewer images are assigned with relatively greater weights. For example, the weight for Spurred Anoda is set to be 18.3 ($1115 / 61$). This strategy, which enforces larger penalties on misclassifications for minority classes, can potentially enhance the classification accuracy for these classes. In preliminary testing, it is observed that the direct inversion of image ratios may lead to sub-optimal performance, hence the final adopted weights are fine-tuned and empirically set as shown in the WCE column in Table~\ref{tab:weights}. Other different choices of weighting strategies are discussed in Section~\ref{sec:weigh_wce}.

\begin{table}[!th]
\caption{Weighting coefficients of weed classes for CE loss, Eqn.~\ref{eqn:weights} and WCE loss functions.}
\centering
\small
\renewcommand{\arraystretch}{1.0}
\resizebox{0.43\textwidth}{!}{
\begin{tabular}{c|c|c|c|c}
\hline
\multirow{2}{*}{} & \multirow{2}{*}{\# of images} & \multicolumn{3}{c}{Weighting Coefficients}                                            \\ \cline{3-5} 
                  &                            & CE & Eqn.~\ref{eqn:weights} & WCE \\ \hline
Morningglory      & 1115                       & 1.0                     & 1.0                                            & 1.4             \\ \hline
Carpetweed        & 763                        & 1.0                     & 1.463                                       & 2.05         \\ \hline
Palmer Amaranth   & 689                        & 1.0                     & 1.62                                       & 2.27         \\ \hline
Waterhemp         & 451                        & 1.0                     & 2.47                                       & 3.46         \\ \hline
Purslane          & 450                        & 1.0                     & 2.48                                       & 3.47         \\ \hline
Nutsedge          & 273                        & 1.0                     & 4.09                                       & 5.72         \\ \hline
Eclipta           & 254                        & 1.0                     & 4.39                                       & 4.39          \\ \hline
Spotted Spurge    & 234                        & 1.0                     & 4.77                                        & 4.77          \\ \hline
Sicklepod         & 240                        & 1.0                     & 4.65                                       & 4.65          \\ \hline
Goosegrass        & 216                        & 1.0                     & 5.16                                        & 5.16           \\ \hline
Prickly Sida      & 129                        & 1.0                     & 8.64                                      & 8.64          \\ \hline
Ragweed           & 129                        & 1.0                     & 8.64                                      & 8.64         \\ \hline
Crabgrass         & 111                        & 1.0                     & 10.05                                       & 10.05          \\ \hline
Swinecress        & 72                         & 1.0                     & 15.49                                     & 15.49         \\ \hline
Spurred Anoda     & 61                         & 1.0                     & 18.28                                      & 18.28         \\ \hline
\end{tabular}}
\label{tab:weights}
\end{table}

\subsection{Deep Learning-based Similarity Measure}
\label{subsec:smi}
To assist in the interpretation of DL classifications,  an inter-class (or within-class) analysis was conducted by quantifying the similarity of the images of weed classes. Euclidean distance is the most commonly used measure of inter-class similarity, but it is sensitive to varying image conditions (e.g., variable ambient light, variations in camera view angle and position), which are typical of the cotton weed images collected under natural field conditions. Cosine similarity (CS), which measures the cosine of the angle between two vectors and is thus not sensitive to magnitude, offers an effective alternative to the Euclidean distance \cite{xia2015learning}.  

In this study, we employed a DL-based CS measure for quantifying inter-class similarities. A DL model was used as feature extractor to obtain hierarchically learnt high-level representation of weed images, based on which  the CS was calculated between two weed classes. Specifically, the VGG11 \cite{simonyan2014very} model was trained on the cotton weed dataset through DTL, and the output of the first FC layer  was taken as the feature vector, which is of length 4,096 (i.e., the output size of the FC layer in the VGG11 network \cite{simonyan2014very}). While other DL models can also be used for feature extraction, the VGG11 was chosen because it achieved the best trade-off between classification performance and training time (see Table~\ref{tab:models}), particularly with high accuracies for minority weed classes (see Table~\ref{tab:wce}). Given the extracted features for any two weed classes, the CS was calculated as follows \cite{xia2015learning} :
\begin{linenomath*}
\begin{equation}\label{eqn:sc}
    CS(x,y) = \frac{1}{N} \sum_i^N \frac{{x_i}^T {y_i}}{\|x_i\| \text{ } \|y_i\|},
\end{equation}
\end{linenomath*}
where $x_i$ and $y_i$ are two feature vectors extracted by the VGG11 model and we randomly sample $N=30$ pairs of images from two weed classes of interest and compute the averaged similarity value. The CS values range from -1 to 1, where 1 means the two classes are perfectly similar and -1 means they are perfectly dissimilar. 

\section{Experimental Results}
\label{sec:results}

\begin{figure*}[!ht]
  \centering
  \includegraphics[width=0.9\textwidth]{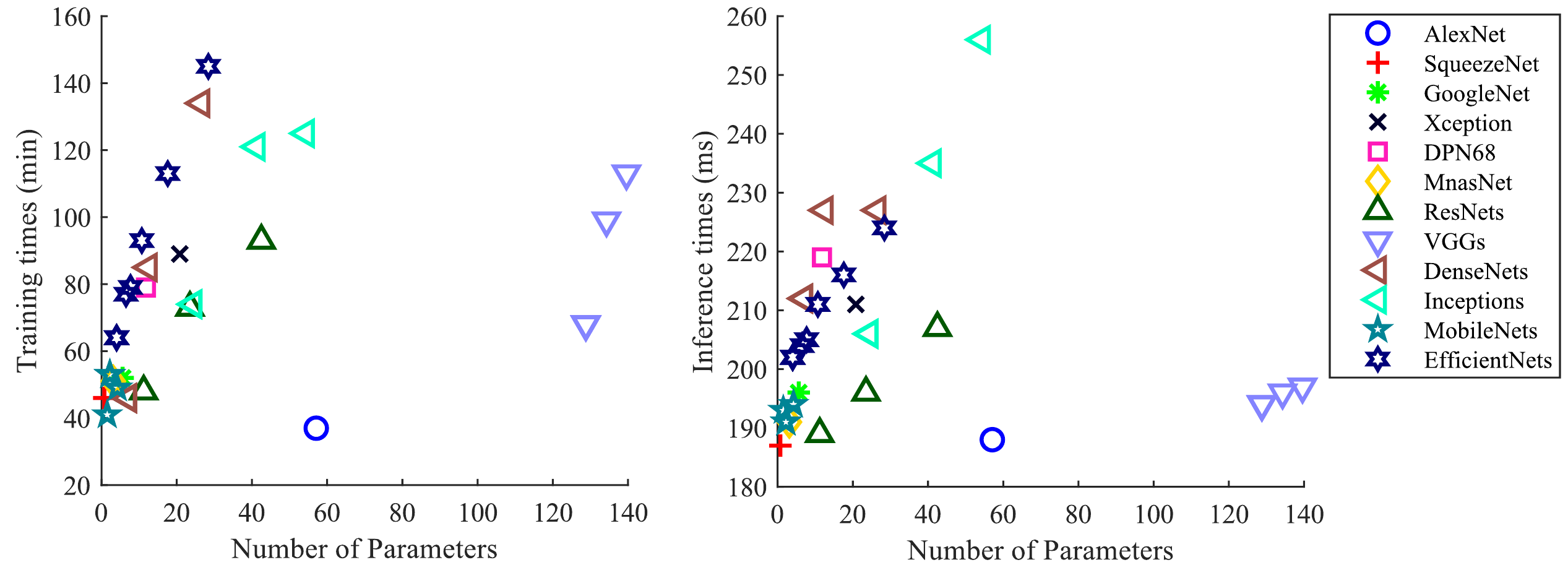}
  \caption{Training and inference time v.s. number of parameters. The DL models are from the same family are labeled with the same marker.}
  \label{fig:time}
\end{figure*}

\begin{figure*}[!ht]
    \centering
    \subfloat{{\includegraphics[width=.475\linewidth]{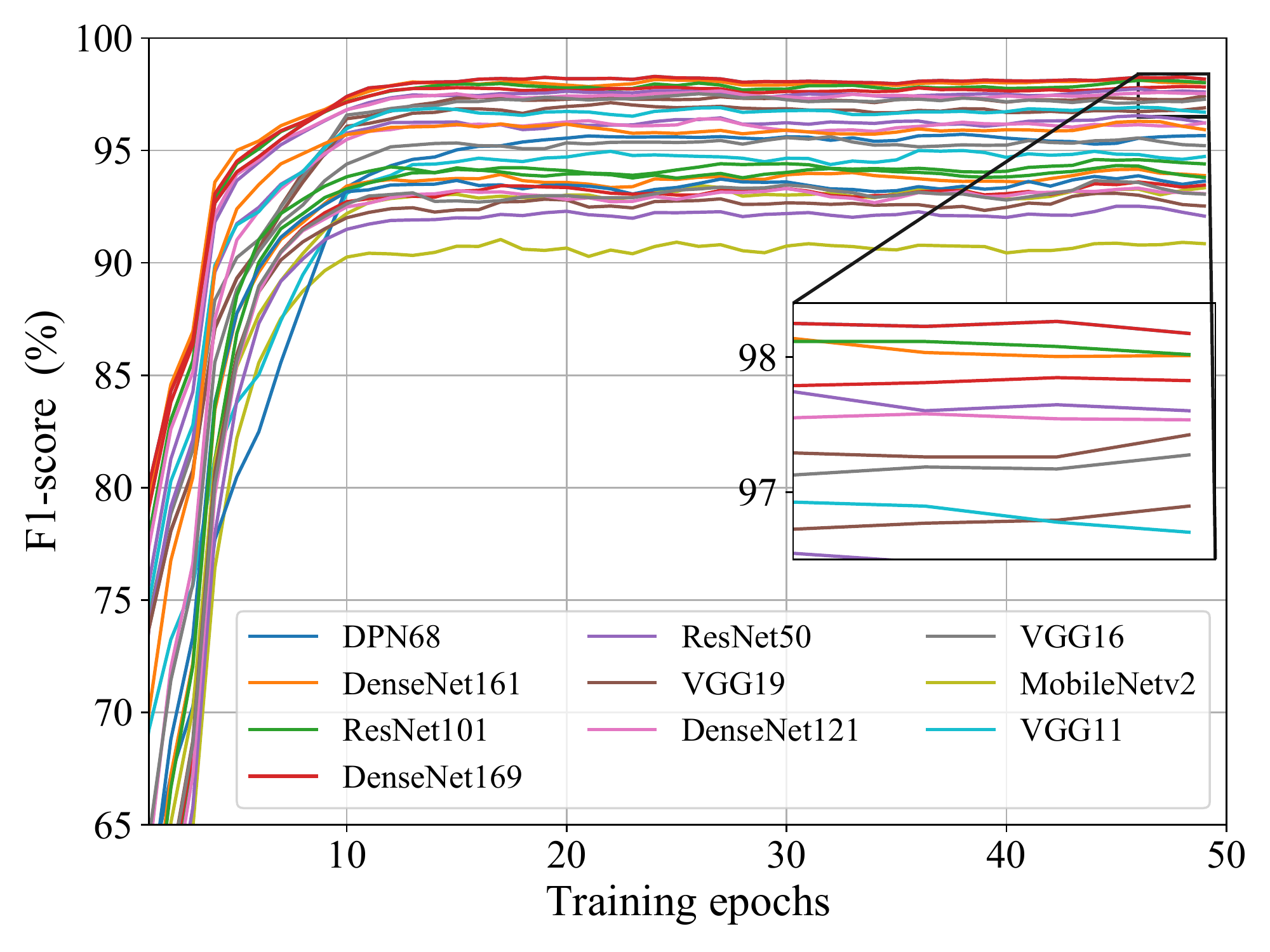} }}%
    \qquad
    \subfloat{{\includegraphics[width=.475\linewidth]{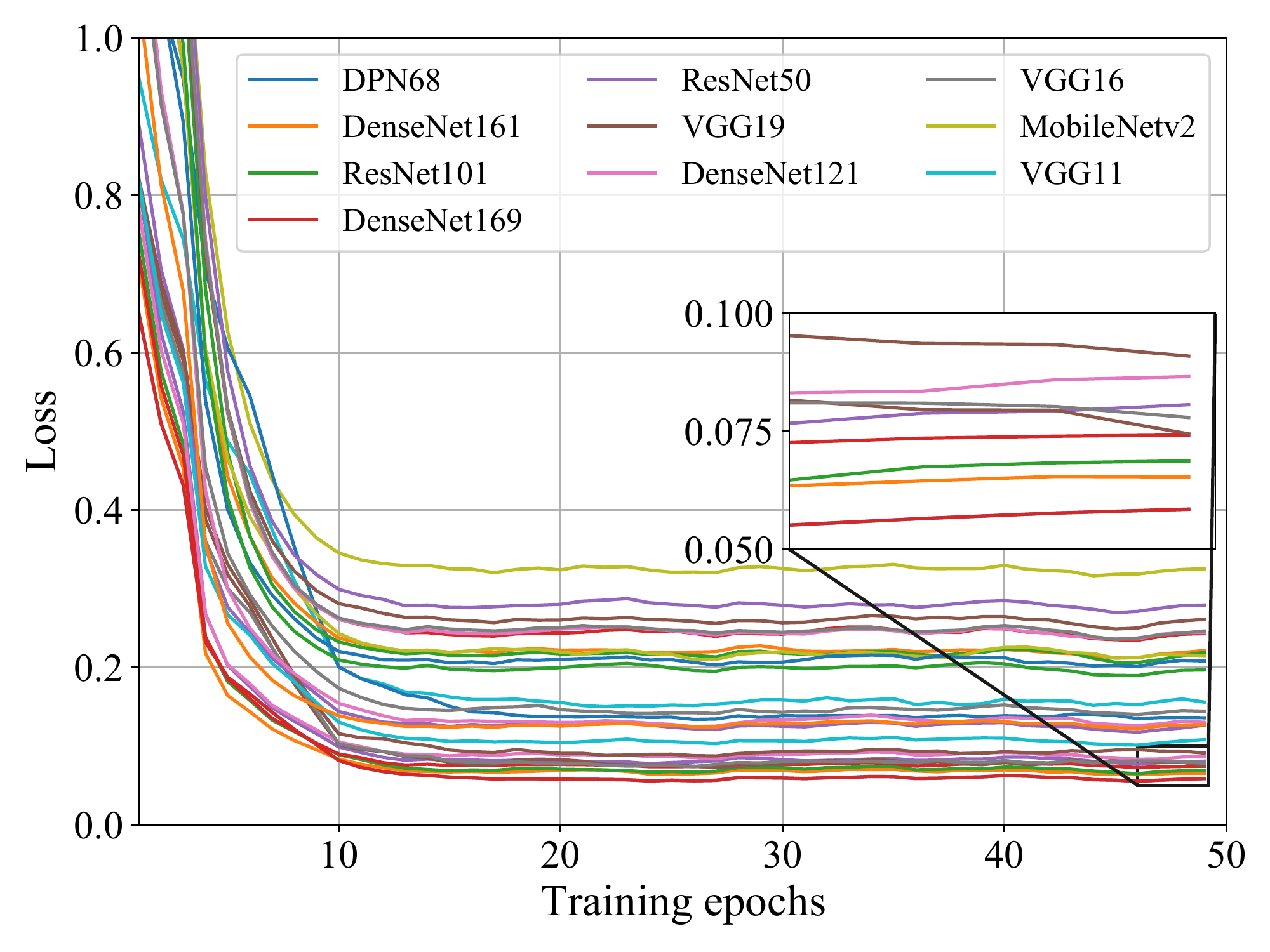}}}
    \caption{Training accuracy and loss curves for the deep learning models.}
    \label{fig:train_acc}
\end{figure*}

\subsection{Deep Learning Model Performance}
\label{subsec:model_performance}
Table~\ref{tab:models} summarizes the number of model parameters, training and inference times and F1 score of the selected 27 DL models. There is a large variation in the number of parameters across the models, ranging from 0.74 M (million) in the SqueezeNet to 139.6 M in the VGG19. Depending on model architectures, the training time ranged from 37 min to 144 min. Models with a larger number of parameters tended to require a longer training time (see Fig.~\ref{fig:time} left), because of increased model complexity. The inference times also exhibited an increasing trend with the number of parameters (see Fig.~\ref{fig:time} right), although it had a notably smaller difference among models, ranging from 188 ms to 256 ms. Inference is mainly a forward propagation process that requires no parameter estimation and is thus far more efficient than training. Particularly, models including AlexNet, SqueezeNet, GoogleNet, ResNet18, ResNet50, VGGs, and MobileNets required inference times less than 200 ms, translating into a prediction speed of over 5 frames per second. The DL models overall show good potential to be deployed for real-time weed identification.

Figure~\ref{fig:train_acc} shows the training accuracy and loss curves of the DL models. All the DL models exhibited promising training performance in terms of fast convergence speed, low training losses and high training accuracies (F1 scores). The training accuracies tended to plateau after 10 epoches at a level exceeding 90\%. 
Regarding test accuracies (Table~\ref{tab:models}), ResNet101 achieved the best overall F1-score of 99.1\%, followed by ResNet50 with the F1=99.0\%. There were other 12 models that gave F1 scores exceeding 98\%, such as three Densenet variants, DPN68, MobilenetV3-large, mong others, and the top-10 models achieved an average F1 score of 98.71\%. On the other hand, three models including AlexNet, SqueezeNet and MnasNet, yieleded the lowest F1 scores that were close to or less than 96\%, although they were all superiorly efficient to train and fast to make inferences.  


\begin{table*}[!ht]
\caption{Performance of 27 state-of-the-art deep learning models on the cotton weed dataset. Note that the variations in training time are negligible so its standard deviation is not included. The top-10 testing accuracies of DL models are highlighted in bold. ``M'' stands for a million.}
\renewcommand{\arraystretch}{1.6}
\centering
\resizebox{0.96\textwidth}{!}{
\begin{tabular}{c|c|c|c|c|c|c|c}
\hline
Index               & \multicolumn{2}{c|}{Model}                                                                            & Parameter Number & Training Time & Training F1-Score & Testing F1-Score    & Inference Time (ms) \\ \hline
1                   & \multicolumn{2}{c|}{AlexNet \cite{krizhevsky2012imagenet}}                                           & 57.1M                      & 37m 2s        & 95.4 ± 0.2        & 95.3 ± 0.4          & 188.5 ± 2.2         \\ \hline
2                   & \multicolumn{2}{c|}{SqueezeNet \cite{iandola2016squeezenet}}                                         & 0.743M                     & 46m 7s        & 96.4 ± 0.2        & 95.8 ± 0.5          & 187.3 ± 1.6         \\ \hline
3                   & \multicolumn{2}{c|}{GoogleNet \cite{szegedy2015going}}                                               & 5.6M                       & 52m 28s       & 94.7 ± 0          & 97.8 ± 0.3          & 196.3 ± 0.5         \\ \hline
4                   & \multicolumn{2}{c|}{Xception \cite{chollet2017xception}}                                             & 20.8M                      & 89m 9s        & 94.7 ± 0.2        & 97.5 ± 0.4          & 211.3 ± 1.8         \\ \hline
5                   & \multicolumn{2}{c|}{\textbf{DPN68} \cite{chen2017dual}}                                                       & 11.8M                      & 79m 10s       & 98.5 ± 0.1        & \textbf{98.8 ± 0.2} & 219.0 ± 6.9         \\ \hline
6                   & \multicolumn{2}{c|}{MnasNet \cite{tan2019mnasnet}}                                                   & 3.1M                       & 51m 3s        & 91.8 ± 0.2        & 96.0 ± 0.4          & 191.2 ± 2.0         \\ \hline
\multirow{3}{*}{7}  & \multirow{3}{*}{ResNet}       & ResNet18 \cite{he2016deep}                                           & 11.2M                      & 47m 30s       & 96.9 ± 0.1        & 98.1 ± 0.2          & 188.9 ± 0.9         \\ \cline{3-8} 
                    &                               & \textbf{ResNet50} \cite{he2016deep}                                           & 23.5M                      & 73m 17s       & 98.0 ± 0.1        & \textbf{99.0 ± 0.1} & 195.6 ± 0.4         \\ \cline{3-8} 
                    &                               & \textbf{ResNet101} \cite{he2016deep}                                          & 42.5M                      & 92m 55s       & 98.3 ± 0.1        & \textbf{99.1 ± 0.2} & 207.0 ± 0.6         \\ \hline
\multirow{3}{*}{10} & \multirow{3}{*}{VGG}          & VGG11  \cite{simonyan2014very}                                       & 128.8M                     & 67m 46s       & 97.3 ± 0.1        & 98.1 ± 0.2          & 194.1 ± 1.3         \\ \cline{3-8} 
                    &                               & VGG16 \cite{simonyan2014very}                                       & 134.3M                     & 99m 25s       & 97.7 ± 0.2        & 98.1 ± 0.3          & 195.7 ± 1.4         \\ \cline{3-8} 
                    &                               & VGG19  \cite{simonyan2014very}                                       & 139.6M                     & 112m 41s      & 97.9 ± 0.1        & 97.9 ± 0.2          & 197.2 ± 1.4         \\ \hline
\multirow{3}{*}{13} & \multirow{3}{*}{Densenet}     & \textbf{Densenet121} \cite{huang2017densely}                                  & 7.0M                       & 75m 40s       & 97.9 ± 0.1        & \textbf{98.7 ± 0.1} & 212.4 ± 0.8         \\ \cline{3-8} 
                    &                               & \textbf{Densenet161} \cite{huang2017densely}                                  & 26.5M                      & 133m 42s      & 98.4± 0.1         & \textbf{98.9 ± 0.4} & 227.4 ± 0.5         \\ \cline{3-8} 
                    &                               & \textbf{Densenet169} \cite{huang2017densely}                                  & 12.5M                      & 85m 1s        & 98.1 ± 0.1        & \textbf{98.9 ± 0.3} & 226.8 ± 0.5         \\ \hline
\multirow{3}{*}{16} & \multirow{3}{*}{Inception}    & \textbf{Inception v3}  \cite{szegedy2016rethinking}                           & 24.4M                      & 73m 50s       & 96.7 ± 0          & \textbf{98.4 ± 0.3} & 206.3 ± 0.4         \\ \cline{3-8} 
                    &                               & Inception v4 \cite{szegedy2017inception}                             & 41.2M                      & 120m 42s      & 95.9 ± 0.1        & 98.1 ± 0.4          & 235.4 ± 0.8         \\ \cline{3-8} 
                    &                               & \multicolumn{1}{l|}{Inception-ResNet v2 \cite{szegedy2017inception}} & 54.3M                      & 124m 36s      & 94.0 ± 0.2        & 97.6 ± 0.4          & 255.9 ± 1.4         \\ \hline
\multirow{3}{*}{19} & \multirow{3}{*}{Mobilenet}    & \textbf{MobilenetV2} \cite{sandler2018mobilenetv2}                            & 2.2M                       & 53m 27s       & 97.4 ± 0.1        & \textbf{98.4 ± 0.1} & 191.1 ± 0.8         \\ \cline{3-8} 
                    &                               & MobilenetV3-small \cite{howard2019searching}                         & 1.5M                       & 41m 27s       & 94.5 ± 0.2        & 96.6 ± 0.1          & 193.1 ± 1.2         \\ \cline{3-8} 
                    &                               & \textbf{MobilenetV3-large} \cite{howard2019searching}                         & 4.2M                       & 49m 4s        & 96.6 ± 0.1        & \textbf{98.6 ± 0.2} & 193.8 ± 2.0         \\ \hline
\multirow{6}{*}{22} & \multirow{6}{*}{EfficientNet} & EfficientNet-b0 \cite{tan2019efficientnet}                           & 4.0M                       & 63m 39s       & 93.0 ± 0.1        & 97.4 ± 0.4          & 202.0 ± 5.6         \\ \cline{3-8} 
                    &                               & EfficientNet-b1 \cite{tan2019efficientnet}                           & 6.5M                       & 77m 8s        & 93.8 ± 0.2        & 97.3 ± 0.4          & 203.8 ± 0.8         \\ \cline{3-8} 
                    &                               & EfficientNet-b2 \cite{tan2019efficientnet}                           & 7.7M                       & 78m 56s       & 94.1 ± 0.2        & 97.8 ± 0.1          & 204.5 ± 1.7         \\ \cline{3-8} 
                    &                               & \textbf{EfficientNet-b3} \cite{tan2019efficientnet}                           & 10.7M                      & 92m 51s       & 95.0 ± 0.2        & \textbf{98.2 ± 0.1} & 211.3 ± 1.2         \\ \cline{3-8} 
                    &                               & EfficientNet-b4 \cite{tan2019efficientnet}                           & 17.6M                      & 113m 12s      & 94.1 ± 0.2        & 97.8 ± 0.2          & 216.3 ± 1.3         \\ \cline{3-8} 
                    &                               & EfficientNet-b5 \cite{tan2019efficientnet}                           & 28.4M                      & 144m 44s      & 94.1 ± 0.3        & 97.4 ± 0.1          & 224.1 ± 1.5         \\ \hline

\end{tabular}}
\label{tab:models}
\end{table*}

\begin{figure*}[!ht]
\centering
\includegraphics[width=0.99\textwidth]{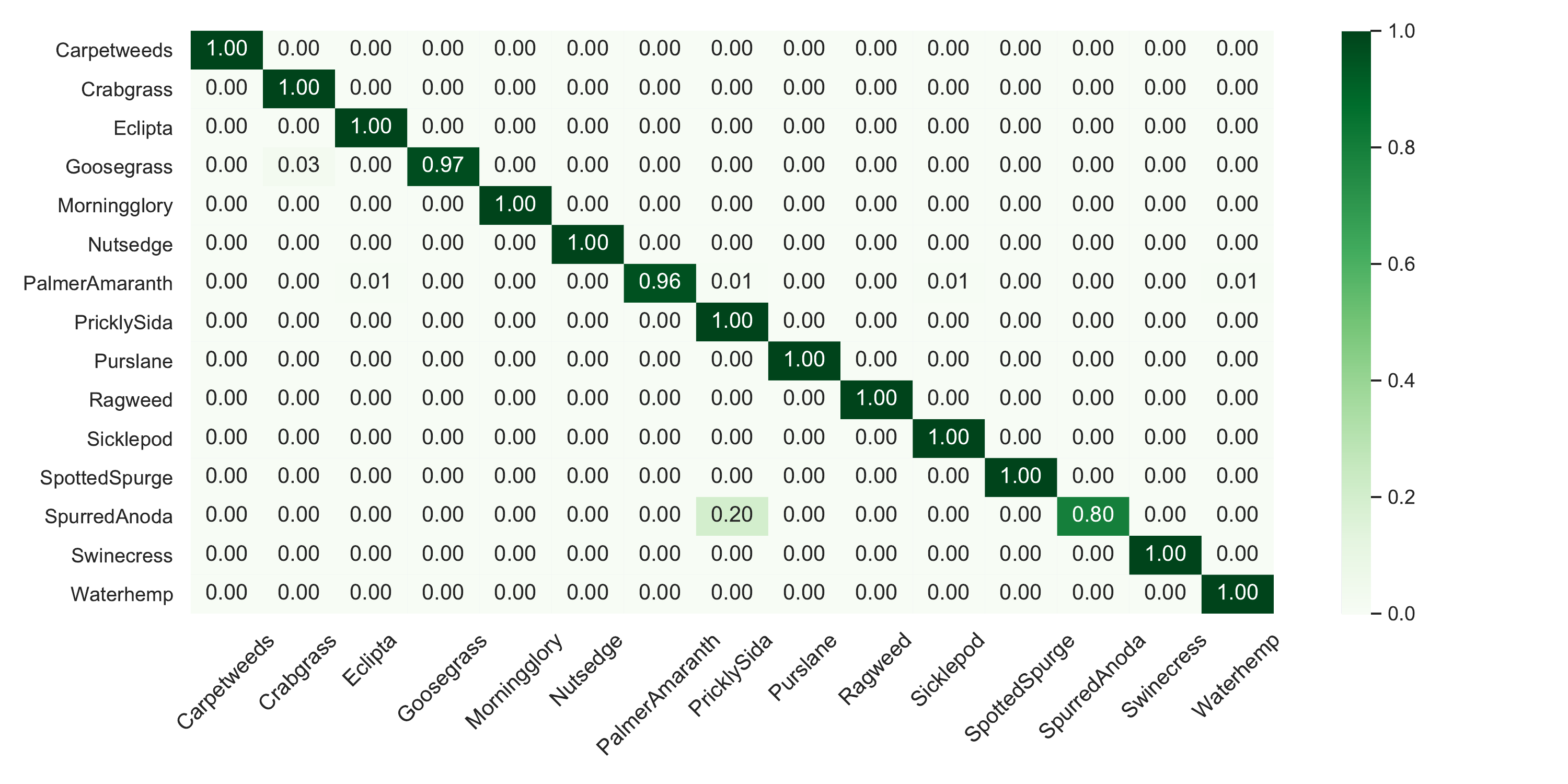}
\caption{The confusion matrix of the ResNet-101 model on the test dataset.}
\label{fig:resnet_cm}
\end{figure*}

\begin{figure*}[!ht]
\centering
\includegraphics[width=0.99\textwidth]{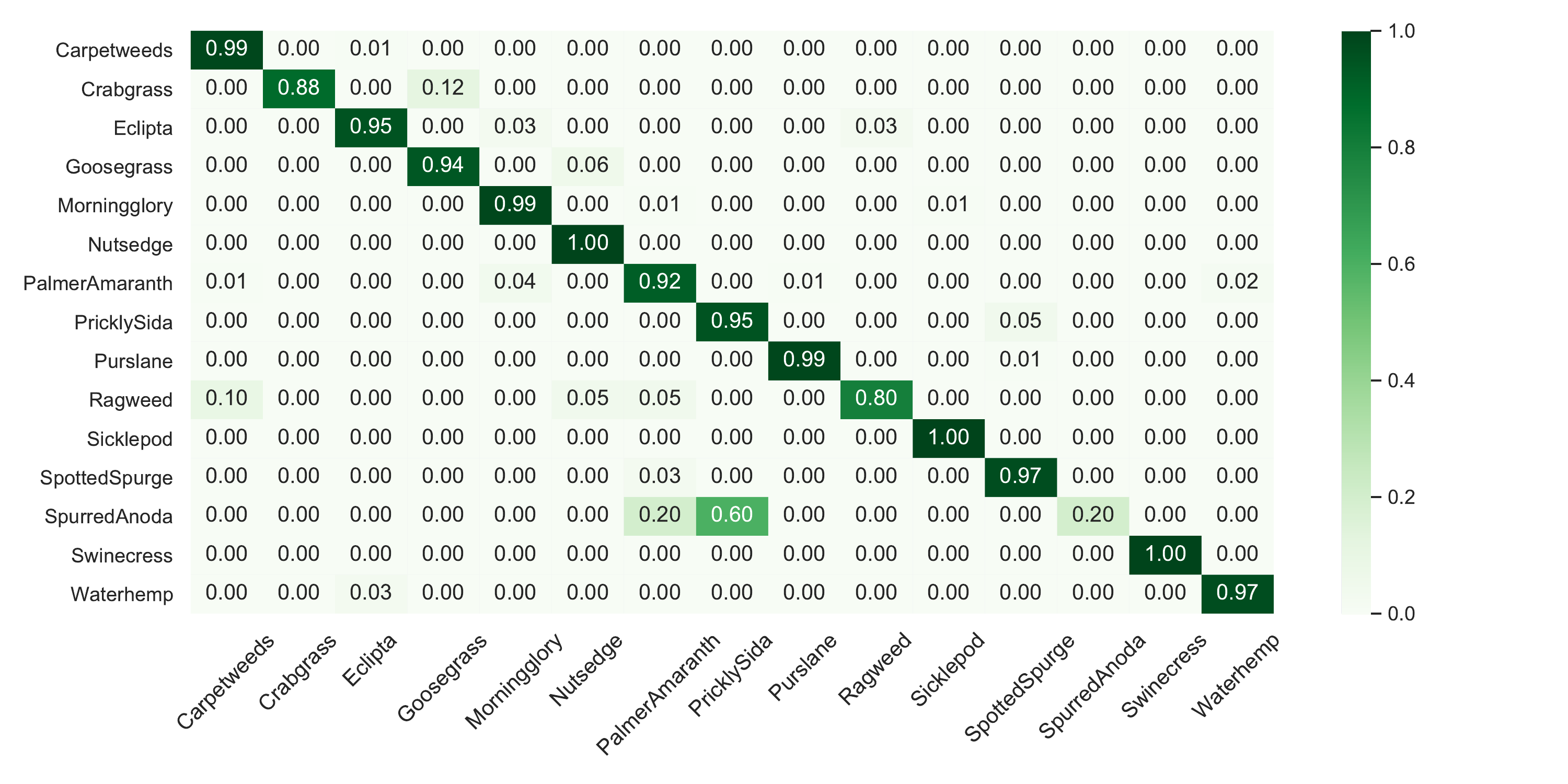}
\caption{The confusion matrix of the MnasNet model on the test dataset.}
\label{fig:mnasnet_cm}
\end{figure*}

The confusion matrices on test data for all the DL models are available on our Github page\footnote{\url{https://github.com/Derekabc/CottonWeeds/tree/master/Confusing_Matrices}}. Due to space constraints, we only show the confusion matrices for one top F1-score model, ResNet-101, and one low-performant model, MnasNet (MnasNet1.0), in Fig.~\ref{fig:resnet_cm} and Fig.~\ref{fig:mnasnet_cm}, respectively. The ResNet-101 yielded perfect classifications for 12 out of 15 weed classes, although it misclassified 3\%, 4\% and 20\% of the images of Goosegrass, Palmer Amaranth and Spurred Annoda, respectively. Spurred Anoda was the most challenging weed class to distinguish from others. The ResNet-101 achieved the classification accuracy of 80\% for this species, misclassifying 20.0\% of the weed as PricklySida. The MnasNet model  only achieved the accuracy of 20\% in the identification of Spurred Annoda, as shown in Fig.~\ref{fig:mnasnet_cm}, misclassifying 60\% and 20\% of the weed as Prickly Sida and Palmer Amaranth, respectively. The poor accuracies are presumably because of the smallest number of images available in the dataset (61 as shown in Fig.~\ref{fig:bar_plot}). Similar low accuracies were also observed for other minority weed classes such as Crabgrass and Ragweed, with an accuracy of 88\% and 80\%, respectively, by the MasNet. To improve the performance of DL models on the minority weed classes, the proposed WCE loss function is discussed next.

\subsection{Performance Improvement with the WCE Loss}
\label{subsec:wce_r}

\begin{figure*}[!ht]
\centering
\includegraphics[width=0.99\textwidth]{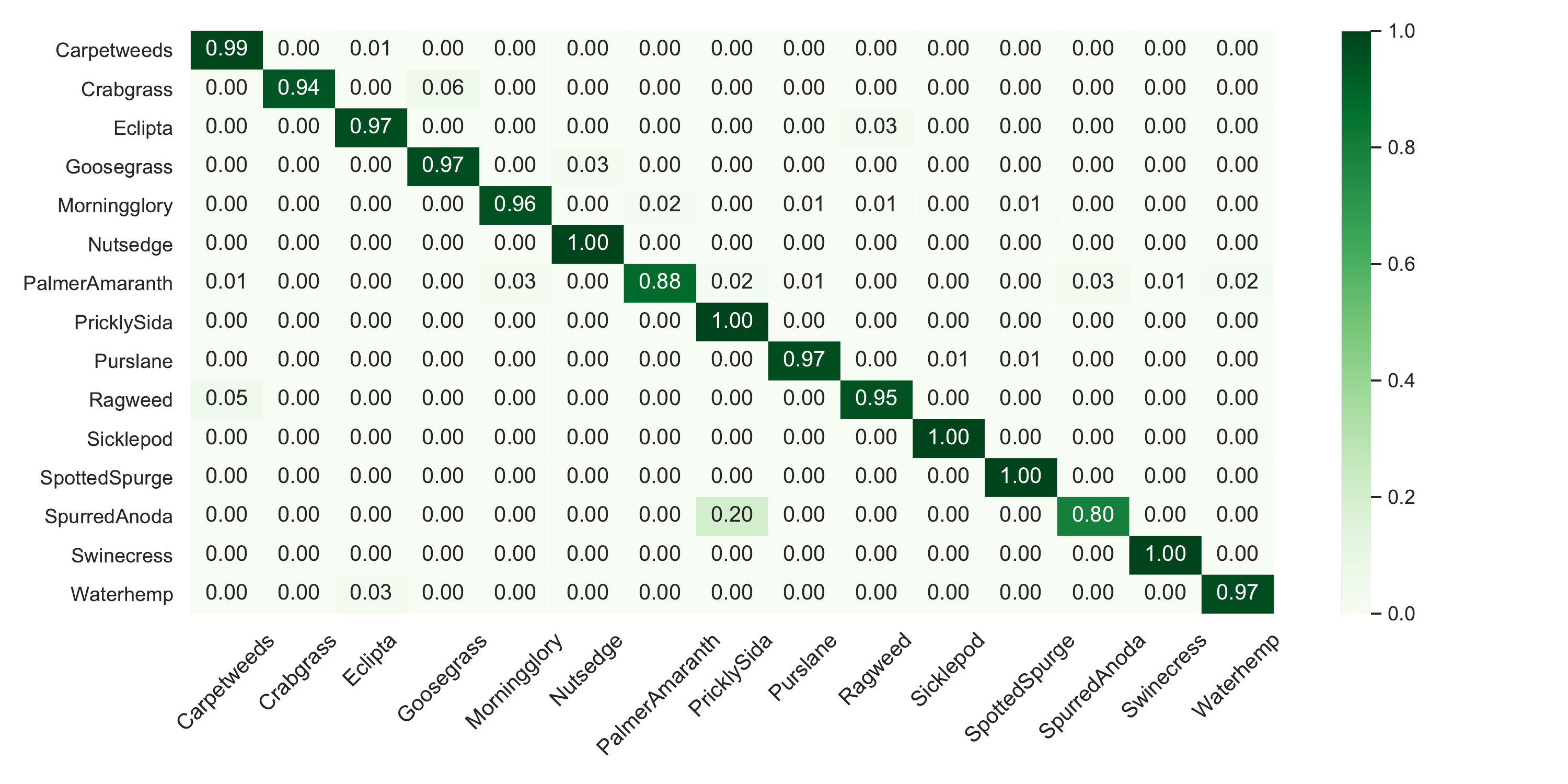}
\caption{The confusion matrix of the MnasNet model with the weighted cross entropy loss strategy evaluated on the test dataset.}
\label{fig:mnasnet_weighted}
\end{figure*}

Fig.~\ref{fig:mnasnet_weighted} shows the confusion matrix achieved by the MnasNet model trained with the WCE loss function (Eqn.~\ref{eqn:wce_loss}). The WCE-based model achieved remarkable improvements over the counterpart (Fig.~\ref{fig:mnasnet_cm}) trained with the regular CE function (Eqn.~\ref{eqn:ce_loss}) in classifying minority weed classes. The classification accuracy of Spurred Annoda jumped from 20\% to 80\%, and the accuracies for Crabgrass and Ragweed were improved from 88\% to 94\% and from 80\% to 95\%, respectively.

Table~\ref{tab:wce} compares the classification accuracies by five selected models (including the aforementioned MnasNet) using the CE loss and the WCE loss. The confusion matrices for all the DL models trained with the CE and WCE loss separately are available at the site\footnote{\url{https://github.com/Derekabc/CottonWeeds/tree/master/Confusing_Matrices}}. Emphasis here is placed on classifying two majority weeds, Morningglory and Waterhemp, and two minority weeds, Crabgrass and Spurred Anoda. Considerable improvements were achieved by all these models for the minority weed classes. Notably, in addition to MnasNet, EfficientNet-b2 and Xception achieved improvements of 40\% and 20\% in identifying Spurred Anoda, respectively, compared to the models the CE loss. 

Despite improvements on minority classes, models like Xception, MnasNet and EfficientNet-b2 resulted in a slightly decreased accuracy for Morningglory. This is because the WCE strategy that placed stronger weights on minority classes might negatively affect the classification for major classes. Nonetheless, the significant improvements on the minority classes outweighed the decrease accuracy in the majority classes, leading to overall improvements in F1-score by these models. Particularly, DenseNet161 achieved an overall F1-score of 99.24\%, outperforming the ResNet101 that achieved the best accuracy (99.1\%) among all the CE-based models. The VGG11 saw a slight decrease in the overall F1-score, but it is encouraging to see that the model achieved 100\% classification for Spurred Anoda that has only 61 images in the weed.

\begin{table*}[!th]
\centering
\renewcommand{\arraystretch}{1.4}
\caption{Performance comparison between cross entropy (CE) loss and weighted cross entropy (WCE) loss for each weed class and the overall F1-score (\%).}
\resizebox{0.68\textwidth}{!}{%
\begin{tabular}{c|c|c|c|c|c|c|c|c|c|c}
\hline
\multirow{2}{*}{} & \multicolumn{2}{c|}{Morningglory} & \multicolumn{2}{c|}{Waterhemp} & \multicolumn{2}{c|}{Crabgrass} & \multicolumn{2}{c|}{Spurred Anoda} & \multicolumn{2}{c}{Overall F1-score} \\ \cline{2-11} 
                  & CE              & WCE             & CE             & WCE           & CE             & WCE           & CE              & WCE              & CE                   & WCE                 \\ \hline
Densenet161       & 100             & 100             & 98.53          & 100           & 100            & 100           & 70              & 80               & 98.85                & 99.24               \\ \hline
Xception          & 100             & 98.21           & 100            & 100           & 94.12          & 100           & 50              & 70               & 97.58                & 97.96               \\ \hline
Mnasnet           & 98.81           & 95.83           & 97.06          & 97.06         & 88.24          & 94.12         & 20              & 80               & 95.67                & 96.06               \\ \hline
EfficientNet-b2   & 98.81           & 97.02           & 100            & 100           & 94.12          & 100           & 40              & 80               & 97.71                & 97.96               \\ \hline
VGG11             & 99.4            & 97.02           & 98.53          & 100           & 100            & 94.12         & 90              & 100              & 97.84                & 97.07               \\ \hline
\end{tabular}
}
\label{tab:wce}
\end{table*}

\subsection{Weed Similarity Analysis}
\label{subsec:smir}
\begin{figure*}[!ht]
\centering
\includegraphics[width=0.99\textwidth]{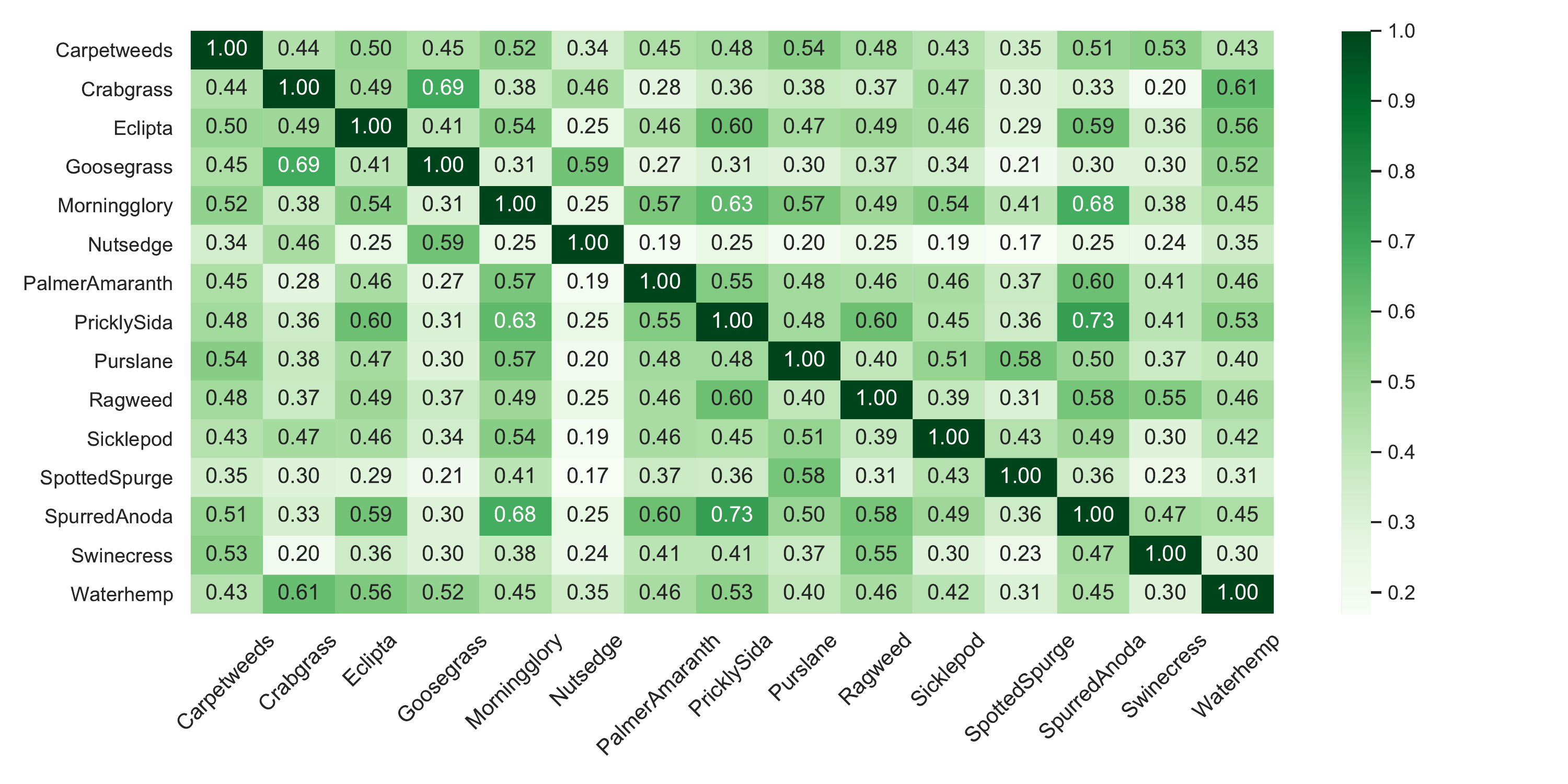}
\caption{The similarity matrix achieved by the deep learning-based similarity measure scheme \ref{subsec:smi} on the cotton weed dataset. The diagonal entries indicate the perfect similarity of each class with itself}
\label{fig:sc}
\end{figure*}

Fig.~\ref{fig:sc} shows an inter-class CS (cosine similarity) matrix based on the features extracted by the VGG11 model (with the CE loss) (Section~\ref{subsec:smi}). The CS matrix helps explain misclassifications by DL models among weed classes. Weed classes that share more common features tended to have higher CS values. For example, Goosegrass and Crabgrass that are both grassy weeds in the Poaceae family, had a CS of 0.69, greater than their similarities with all other weeds. The high CS is in agreement with the classification errors observed between the two classes (see Fig.~\ref{fig:resnet_cm}, Fig.~\ref{fig:mnasnet_cm} and Fig.~\ref{fig:mnasnet_weighted}). For the ResNet101 model, for instance, all the 3\% misclassifications for Goosegrass were due to misclassifying the weed as Crabgrass (Fig.~\ref{fig:resnet_cm}). Spurred Anoda and Prickly Sida are another pair of similar weeds, which both have toothed leaf margins and are members of the Mallow family. The globally highest CS of 0.73 was observed between the two classes. Their strong similarity, along with the fact that Prickly Sida has more than twice as many images as Spurred Anoda, explains the significant proportion of Spurred Anoda misclassified as Prickly Sida (see Fig.~\ref{fig:resnet_cm}, Fig.~\ref{fig:mnasnet_cm} and Fig.~\ref{fig:mnasnet_weighted}).

\section{Discussion and Future Research}
\label{sec:discussion}
In this section, we  discuss two potential approaches to improving the performance on minority weed classes, which will be investigated in future studies.




\subsection{Weighted Loss Functions}
\label{sec:weigh_wce}
The WCE loss function (Eqn.~\ref{eqn:wce_loss}) improves the CE loss by adaptively assigning weights to individual weed classes to account for class imbalance. In addition to the weighting in Eqn.~\ref{eqn:weights}, there are other weighting \cite{phan2020resolving} or cost-sensitive methods \cite{khan2017cost} to cope with imbalanced data. 

The class-balanced (CB) loss introduced in \cite{cui2019class} re-balances the classification loss based on the effective number of samples for each class. 

The CB loss is defined as:
\begin{linenomath*}
\begin{equation}
    CB_{loss} = - \sum_{i=1}^C \frac{1 - \beta}{1 - \beta^n}\text{ }t_i \text{ } \log(p_i),
\label{eqn:es}
\end{equation}
\end{linenomath*}
When $\beta = 0$, the CB loss is equivalent to the CE loss and $\beta \rightarrow 1$ corresponds to re-weighting by inverse class frequency, which enables us to smoothly adjust the class-balanced term between no re-weighting and re-weighing by inverse class frequency \cite{cui2019class}.

Focal loss (FL), which was originally proposed in \cite{lin2017focal}, offers another promising alternative for imbalanced learning, which is calculated as follows:
\begin{linenomath*}
\begin{equation}
    FL_{loss} = - \sum_{i=1}^C (1 - p_i)^{\gamma}\text{ } t_i \log(p_i),
\label{eqn:log}
\end{equation}
\end{linenomath*}
where $(1 - p_i)^{\gamma}$ is called the modulating factor, which allows down-weighting the contributions of easy examples or majority classes during training while rapidly focusing  on challenging classes that have few images. 
Here, $\gamma \geq 0$ is the focusing parameter, and FL loss is reduced to the conventional CE loss when equivalent to $\gamma = 0$.

In future research, we will experiment and evaluate these weighted loss functions for improved classification of minority weed classes.

\subsection{Data Augmentation}
In this paper, although overall DL models achieved remarkable weed identification accuracy, some models that prove to be powerful in visual categorization tasks, like EfficientNet \cite{tan2019efficientnet}, did not perform well as expected, especially on minority weed classes\footnote{\url{https://github.com/Derekabc/CottonWeeds/tree/master/Confusing_Matrices/csv}}. This is likely due to the fact that these models are heavily reliant on large-scale data to be sufficiently optimized while avoiding overfitting \cite{shorten2019survey}. One intuitive solution is to collect more images for the under-performed weed classes. Unfortunately, many weed species may be difficult to collect due to unpredicted weather conditions and limited access to a diversity of field sites. 

Data augmentation (DA) offers an effective means to address the insufficiency of physically collected image data. In DA, a suite of techniques \cite{shorten2019survey}, such as geometric transformations, color space augmentations and generative adversarial networks (GANs), can be used to enhances the size and quality of training images such that deep learning models can be trained on the artificially expanded dataset and then gain better performance. Particularly, GANs have received increasing attention, representing a novel framework of generative modeling through adversarial training \cite{creswell2018generative}. Recently, GAN methods have been investigated for weed classification tasks \cite{espejo2021combining, espejo2021combining} to address the lack of large-scale domain datasets. In this paper, we also applied geometric transformation methods such as random rotation and pixel normalization, but did not fully explore the potential of DA techniques in the classification of weed images, which will be a subject of future research.

\section{Conclusion}
\label{sec:conclu}
In this study, a first, comprehensive benchmark of a suite of 27 DL models was established through transfer learning for multi-class identification of common weeds specific to cotton production systems. A dedicated dataset consisting of 5187 images of 15 weed classes was created by collecting images under natural light conditions and at varied weed growth stages from a diversity of cotton fields in southern U.S. states over two growth seasons. DTL proved to be effective for achieving high weed classification accuracies (F1 score \textgreater 95\%) within reasonably short training time (\textless 2.5 hours). ResNet101 was the best-performant model in terms of the highest accuracy of F1=99.1\%, and the top-10 models resulted in an average F1-score of 98.71\%. A WCE loss function was proposed for model training, in which individualized weights were assigned to weed classes to account for class imbalance, which achieved substantial improvements in classifying minority weed classes. A DL-based cosine similarity metric was found to be useful for assisting the interpretation of the misclassifications. Both the source codes for model development and evaluation and the weed dataset were made publicly accessible for the research community. This study provides a good foundation for informed choice of DL models for weed classification tasks, and can be beneficial for precision agriculture research at large.

\section*{Authorship Contribution}
\textbf{Dong Chen}: Formal analysis, Software, Writing - original draft; \textbf{Yuzhen Lu}: Conceptualization, Investigation, Data curation, Supervision, Writing - review \& editing; \textbf{Zhaojiang Li}: Resources, Writing - review \& editing; \textbf{Sierra Young}: Data curation, Writing - review \& editing. 

\section*{Acknowledgement}
This work was supported in part by Cotton Incorporated award 21-005. The authors thank Dr. Camp Hand and Dr. Edward Barnes for contributing weed images and Dr. Charlie Cahoon for the assistance in weed identification. We also thank Mr. Shea Hoffman and Mr. Vinay Kumar for helping label the weed images. 

\typeout{}
\bibliographystyle{model1-num-names}
\bibliography{ref}
\end{document}


\title{Supplementary Materials: Performance Evaluation of Transfer Learning on Multiclass Weed Species Identification in Cotton Production Systems} 
\author{Jack, Jack, Jack, Jack}
\date{\today} 
\maketitle 

\tableofcontents




\section{Pseudo-code of Transfer Learning}
Following figures are the pseudo-code of applying deep learning models with transfer learning. For detailed implementations, please refer to the implementation at: \url{https://github.com/Derekabc/CottonWeeds}. Note that different deep learning models have different network structures, thus modification for different models varies.

\begin{figure}[!ht]
\begin{algorithmic}[1]
	\If{\textit{``model'' is AlexNet}}
       \State \textit{model = alexnet(pretrained=True)}
       \State \textit{num\_ftrs = model.fc.in\_features}
       \State \textit{model.classifier[6] = nn.Linear(4096, num\_classes)}
	\EndIf
\end{algorithmic} 
\caption{Transfer learning on AlexNet} 
\end{figure}

\begin{figure}[!ht]
\begin{algorithmic}[1]
	\If{\textit{``model'' is SqueezeNet}}
       \State \textit{model = SqueezeNet(pretrained=True)}
       \State \textit{model.classifier[1] = Conv2d(512, num\_classes)}
	\EndIf
\end{algorithmic} 
\caption{Transfer learning on SqueezeNet} 
\end{figure}

\begin{figure}[!ht]
\begin{algorithmic}[1]
	\If{\textit{``model'' is GoogleNet	}}
       \State \textit{model = GoogleNet(pretrained=True)}
       \State \textit{num\_ftrs = model.fc.in\_features}
       \State \textit{model.fc = nn.Linear(num\_ftrs, num\_classes)}
	\EndIf
\end{algorithmic} 
\caption{Transfer learning on GoogleNet} 
\end{figure}

\begin{figure}[!ht]
\begin{algorithmic}[1]
	\If{\textit{``model'' is Xception}}
       \State \textit{model = Xception(pretrained=True)}
       \State \textit{num\_ftrs = model.last\_linear.in\_features}
       \State \textit{model.last\_linear = nn.Linear(num\_ftrs, num\_classes)}
	\EndIf
\end{algorithmic} 
\caption{Transfer learning on Xception} 
\end{figure}

\begin{figure}[!ht]
\begin{algorithmic}[1]
	\If{\textit{``model'' is DPN68}}
      \State \textit{model = DPN68(pretrained=True)}
      \State \textit{model.last\_linear = Conv2d(832, num\_classes)}
	\EndIf
\end{algorithmic} 
\caption{Transfer learning on SqueezeNet} 
\end{figure}

\begin{figure}[!ht]
\begin{algorithmic}[1]
	\If{\textit{``model'' is MnasNet}}
      \State \textit{model = MnasNet(pretrained=True)}
      \State \textit{num\_ftrs = model.classifier[1].in\_features}
      \State \textit{model.last\_linear = nn.Linear(num\_ftrs, num\_classes)}
	\EndIf
\end{algorithmic} 
\caption{Transfer learning on MnasNet} 
\end{figure}

\begin{figure}[!ht]
\begin{algorithmic}[1]
	\If{\textit{``model'' is ResNet18}}
      \State \textit{model = resnet18(pretrained=True)}
      \State \textit{num\_ftrs = model.fc.in\_features}
      \State \textit{model.fc = FC(num\_ftrs, num\_classes)}
	\EndIf
\end{algorithmic} 
\caption{Transfer learning on ResNet18} 
\end{figure}

\begin{figure}[!ht]
\begin{algorithmic}[1]
	\If{\textit{``model'' is ResNet50}}
      \State \textit{model = resnet50(pretrained=True)}
      \State \textit{num\_ftrs = model.fc.in\_features}
      \State \textit{model.fc = FC(num\_ftrs, num\_classes)}
	\EndIf
\end{algorithmic} 
\caption{Transfer learning on ResNet50} 
\end{figure}

\begin{figure}[!ht]
\begin{algorithmic}[1]
	\If{\textit{``model'' is ResNet101}}
      \State \textit{model = resnet101(pretrained=True)}
      \State \textit{num\_ftrs = model.fc.in\_features}
      \State \textit{model.fc = FC(num\_ftrs, num\_classes)}
	\EndIf
\end{algorithmic} 
\caption{Transfer learning on ResNet101} 
\end{figure}

\begin{figure}[!ht]
\begin{algorithmic}[1]
	\If{\textit{``model'' is VGG11}}
      \State \textit{model = VGG11(pretrained=True)}
      \State \textit{model.classifier[6] = FC(4096, num\_classes)}
	\EndIf
\end{algorithmic} 
\caption{Transfer learning on VGG11} 
\end{figure}

\begin{figure}[!ht]
\begin{algorithmic}[1]
	\If{\textit{``model'' is VGG16}}
      \State \textit{model = VGG16(pretrained=True)}
      \State \textit{model.classifier[6] = FC(4096, num\_classes)}
	\EndIf
\end{algorithmic} 
\caption{Transfer learning on VGG16} 
\end{figure}

\begin{figure}[!ht]
\begin{algorithmic}[1]
	\If{\textit{``model'' is VGG19}}
      \State \textit{model = VGG19(pretrained=True)}
      \State \textit{model.classifier[6] = FC(4096, num\_classes)}
	\EndIf
\end{algorithmic} 
\caption{Transfer learning on VGG19} 
\end{figure}

\begin{figure}[!ht]
\begin{algorithmic}[1]
	\If{\textit{``model'' is DenseNet121}}
      \State \textit{model = DenseNet121(pretrained=True)}
      \State \textit{num\_ftrs = model.classifier.in\_features}
      \State \textit{model.classifier  = FC(num\_ftrs, num\_classes)}
	\EndIf
\end{algorithmic} 
\caption{Transfer learning on DenseNet121} 
\end{figure}

\begin{figure}[!ht]
\begin{algorithmic}[1]
	\If{\textit{``model'' is DenseNet161}}
      \State \textit{model = DenseNet161(pretrained=True)}
      \State \textit{num\_ftrs = model.classifier.in\_features}
      \State \textit{model.classifier  = FC(num\_ftrs, num\_classes)}
	\EndIf
\end{algorithmic} 
\caption{Transfer learning on DenseNet161} 
\end{figure}

\begin{figure}[!ht]
\begin{algorithmic}[1]
	\If{\textit{``model'' is DenseNet169}}
      \State \textit{model = DenseNet169(pretrained=True)}
      \State \textit{num\_ftrs = model.classifier.in\_features}
      \State \textit{model.classifier  = FC(num\_ftrs, num\_classes)}
	\EndIf
\end{algorithmic} 
\caption{Transfer learning on DenseNet169} 
\end{figure}

\begin{figure}[!ht]
\begin{algorithmic}[1]
	\If{\textit{``model'' is Inception-v3}}
      \State \textit{model = Inception-v3(pretrained=True)}
      \State \textit{model\_ft.aux\_logits = False)}
      \State \textit{num\_ftrs = model.fc.in\_features}
      \State \textit{model.fc  = FC(num\_ftrs, num\_classes)}
	\EndIf
\end{algorithmic} 
\caption{Transfer learning on Inception-v3} 
\end{figure}

\begin{figure}[!ht]
\begin{algorithmic}[1]
	\If{\textit{``model'' is Inception-v4}}
      \State \textit{model = Inception-v4(pretrained=True)}
      \State \textit{num\_ftrs = model.last\_linear .in\_features}
      \State \textit{model.last\_linear  = FC(num\_ftrs, num\_classes)}
	\EndIf
\end{algorithmic} 
\caption{Transfer learning on Inception-v4} 
\end{figure}

\begin{figure}[!ht]
\begin{algorithmic}[1]
	\If{\textit{``model'' is Inception-ResNet}}
      \State \textit{model = Inception-ResNet(pretrained=True)}
      \State \textit{num\_ftrs = model.last\_linear .in\_features}
      \State \textit{model.last\_linear  = FC(num\_ftrs, num\_classes)}
	\EndIf
\end{algorithmic} 
\caption{Transfer learning on Inception-ResNet} 
\end{figure}

\begin{figure}[!ht]
\begin{algorithmic}[1]
	\If{\textit{``model'' is MobileNet-v2}}
      \State \textit{model = MobileNet-v2(pretrained=True)}
      \State \textit{model.classifier[1]  = FC(model.last\_channel, num\_classes)}
	\EndIf
\end{algorithmic} 
\caption{Transfer learning on MobileNet-v2} 
\end{figure}

\begin{figure}[!ht]
\begin{algorithmic}[1]
	\If{\textit{``model'' is MobileNet-v3-small}}
      \State \textit{model = MobileNet-v3-small(pretrained=True)}
      \State \textit{model.classifier[3]  = FC(model.classifier[3].in\_features, num\_classes)}
	\EndIf
\end{algorithmic} 
\caption{Transfer learning on MobileNet-v3-small} 
\end{figure}

\begin{figure}[!ht]
\begin{algorithmic}[1]
	\If{\textit{``model'' is MobileNet-v3-large}}
      \State \textit{model = MobileNet-v3-large(pretrained=True)}
      \State \textit{model.classifier[3]  = FC(model.classifier[3].in\_features, num\_classes)}
	\EndIf
\end{algorithmic} 
\caption{Transfer learning on MobileNet-v3-large} 
\end{figure}

\begin{figure}[!ht]
\begin{algorithmic}[1]
	\If{\textit{``model'' is EfficientNet-b0}}
      \State \textit{model = EfficientNet-b0(pretrained=True)}
      \State \textit{model.classifier[1]  = FC(model.last\_channel, num\_classes)}
	\EndIf
\end{algorithmic} 
\caption{Transfer learning on EfficientNet-b0} 
\end{figure}

\begin{figure}[!ht]
\begin{algorithmic}[1]
	\If{\textit{``model'' is EfficientNet-b1}}
      \State \textit{model = EfficientNet-b1(pretrained=True)}
      \State \textit{model.classifier[1]  = FC(model.last\_channel, num\_classes)}
	\EndIf
\end{algorithmic} 
\caption{Transfer learning on EfficientNet-b1} 
\end{figure}

\begin{figure}[!ht]
\begin{algorithmic}[1]
	\If{\textit{``model'' is EfficientNet-b2}}
      \State \textit{model = EfficientNet-b2(pretrained=True)}
      \State \textit{model.classifier[1]  = FC(model.last\_channel, num\_classes)}
	\EndIf
\end{algorithmic} 
\caption{Transfer learning on EfficientNet-b2} 
\end{figure}

\begin{figure}[!ht]
\begin{algorithmic}[1]
	\If{\textit{``model'' is EfficientNet-b3}}
      \State \textit{model = EfficientNet-b3(pretrained=True)}
      \State \textit{model.classifier[1]  = FC(model.last\_channel, num\_classes)}
	\EndIf
\end{algorithmic} 
\caption{Transfer learning on EfficientNet-b3} 
\end{figure}

\begin{figure}[!ht]
\begin{algorithmic}[1]
	\If{\textit{``model'' is EfficientNet-b4}}
      \State \textit{model = EfficientNet-b4(pretrained=True)}
      \State \textit{model.classifier[1]  = FC(model.last\_channel, num\_classes)}
	\EndIf
\end{algorithmic} 
\caption{Transfer learning on EfficientNet-b4} 
\end{figure}

\begin{figure}[!ht]
\begin{algorithmic}[1]
	\If{\textit{``model'' is EfficientNet-b5}}
      \State \textit{model = EfficientNet-b5(pretrained=True)}
      \State \textit{model.classifier[1]  = FC(model.last\_channel, num\_classes)}
	\EndIf
\end{algorithmic} 
\caption{Transfer learning on EfficientNet-b5} 
\end{figure}

\subsection{Weighting Coefficients for the Weighted Cross Entropy}

\begin{sidewaystable}
\normalsize
\renewcommand{\arraystretch}{1.4}
\centering
\resizebox{1.0\textwidth}{!}{
\begin{tabular}{c|c|c|c|c|c|c|c|c|c|c|c|c|c|c|c}
\hline
 & Carpetweeds & Crabgrass & Eclipta & Goosegrass & Morningglory & Nutsedge & PalmerAmaranth & PricklySida & Purslane & Ragweed  & Sicklepod & SpottedSpurge & SpurredAnoda & Swinecress & Waterhemp \\ \hline
Original weights & 1.46      & 10.05    & 4.39  & 5.16      & 1.00            & 4.08   & 1.62         & 8.64      & 2.48   & 8.64   & 4.65    & 4.77         & 18.28      & 15.49    & 2.47    \\ \hline
Updated weights     & 2.05     & 10.05    & 4.39  & 7.23     & 1.40          & 5.72  & 2.27        & 8.64      & 3.47  & 12.10 & 4.65    & 4.77         & 18.28      & 15.49    & 3.46   \\ \hline
\end{tabular}}
\caption{Weighting coefficients for weed classes in the weighted cross entropy loss function.}
\label{tab:weights}
\end{sidewaystable}

\section{Confusion Matrices of Deep learning models (CE Loss)}
Following the confusion matrices achieved by the deep learning models.

\begin{table*}[!th]
\centering
\renewcommand{\arraystretch}{1.5}
\resizebox{1.0\textwidth}{!}{
\begin{tabular}{c|c|c|c|c|c|c|c|c|c|c|c|c|c|c|c}
\hline
\multicolumn{1}{l|}{} & Carpetweeds   & Crabgrass     & Eclipta       & Goosegrass    & Morningglory  & Nutsedge      & PalmerAmaranth & PricklySida   & Purslane      & Ragweed       & Sicklepod     & SpottedSpurge & SpurredAnoda  & Swinecress    & Waterhemp     \\ \hline
Carpetweeds            & \textbf{0.97} & 0.00          & 0.01          & 0.00          & 0.02          & 0.00          & 0.00           & 0.00          & 0.00          & 0.00          & 0.00          & 0.00          & 0.00          & 0.00          & 0.00          \\ \hline
Crabgrass              & 0.00          & \textbf{0.94} & 0.00          & 0.06          & 0.00          & 0.00          & 0.00           & 0.00          & 0.00          & 0.00          & 0.00          & 0.00          & 0.00          & 0.00          & 0.00          \\ \hline
Eclipta                & 0.08          & 0.00          & \textbf{0.85} & 0.00          & 0.00          & 0.00          & 0.00           & 0.03          & 0.00          & 0.00          & 0.00          & 0.00          & 0.03          & 0.00          & 0.03          \\ \hline
Goosegrass             & 0.00          & 0.03          & 0.00          & \textbf{0.97} & 0.00          & 0.00          & 0.00           & 0.00          & 0.00          & 0.00          & 0.00          & 0.00          & 0.00          & 0.00          & 0.00          \\ \hline
Morningglory           & 0.00          & 0.00          & 0.00          & 0.00          & \textbf{0.99} & 0.00          & 0.00           & 0.00          & 0.01          & 0.00          & 0.00          & 0.00          & 0.00          & 0.00          & 0.00          \\ \hline
Nutsedge               & 0.00          & 0.00          & 0.00          & 0.00          & 0.00          & \textbf{1.00} & 0.00           & 0.00          & 0.00          & 0.00          & 0.00          & 0.00          & 0.00          & 0.00          & 0.00          \\
\hline
\end{tabular}}
\caption{The confusion matrix achieved by the Alexnet model on the test subset of the CottonWeeds dataset.}
\label{tab:alextnet}
\end{table*}

\begin{table*}[!th]
\centering
\renewcommand{\arraystretch}{1.5}
\resizebox{1.0\textwidth}{!}{
\begin{tabular}{c|c|c|c|c|c|c|c|c|c|c|c|c|c|c|c}
\hline
\multicolumn{1}{l|}{} & Carpetweeds   & Crabgrass     & Eclipta       & Goosegrass    & Morningglory  & Nutsedge      & PalmerAmaranth & PricklySida   & Purslane      & Ragweed       & Sicklepod     & SpottedSpurge & SpurredAnoda  & Swinecress    & Waterhemp     \\ \hline
Carpetweeds            & \textbf{0.97} & 0.00          & 0.01          & 0.00          & 0.02          & 0.00          & 0.00           & 0.00          & 0.00          & 0.00          & 0.00          & 0.00          & 0.00          & 0.00          & 0.00          \\ \hline
Crabgrass              & 0.00          & \textbf{0.94} & 0.00          & 0.06          & 0.00          & 0.00          & 0.00           & 0.00          & 0.00          & 0.00          & 0.00          & 0.00          & 0.00          & 0.00          & 0.00          \\ \hline
Eclipta                & 0.08          & 0.00          & \textbf{0.85} & 0.00          & 0.00          & 0.00          & 0.00           & 0.03          & 0.00          & 0.00          & 0.00          & 0.00          & 0.03          & 0.00          & 0.03          \\ \hline
Goosegrass             & 0.00          & 0.03          & 0.00          & \textbf{0.97} & 0.00          & 0.00          & 0.00           & 0.00          & 0.00          & 0.00          & 0.00          & 0.00          & 0.00          & 0.00          & 0.00          \\ \hline
Morningglory           & 0.00          & 0.00          & 0.00          & 0.00          & \textbf{0.99} & 0.00          & 0.00           & 0.00          & 0.01          & 0.00          & 0.00          & 0.00          & 0.00          & 0.00          & 0.00          \\ \hline
Nutsedge               & 0.00          & 0.00          & 0.00          & 0.00          & 0.00          & \textbf{1.00} & 0.00           & 0.00          & 0.00          & 0.00          & 0.00          & 0.00          & 0.00          & 0.00          & 0.00          \\
\hline

\end{tabular}}
\caption{The confusion matrix achieved by the Squeezenet model on the test subset of the CottonWeeds dataset.}
\label{tab:Squeezenet}
\end{table*}

\begin{table*}[!th]
\centering
\renewcommand{\arraystretch}{1.5}
\resizebox{1.0\textwidth}{!}{
\begin{tabular}{c|c|c|c|c|c|c|c|c|c|c|c|c|c|c|c}
\hline
\multicolumn{1}{l|}{} & Carpetweeds   & Crabgrass     & Eclipta       & Goosegrass    & Morningglory  & Nutsedge      & PalmerAmaranth & PricklySida   & Purslane      & Ragweed       & Sicklepod     & SpottedSpurge & SpurredAnoda  & Swinecress    & Waterhemp     \\ \hline
Carpetweeds            & \textbf{0.99} & 0.00          & 0.01          & 0.00          & 0.00          & 0.00          & 0.00           & 0.00          & 0.00          & 0.00          & 0.00          & 0.00          & 0.00          & 0.00          & 0.00          \\ \hline
Crabgrass              & 0.00          & \textbf{0.94} & 0.00          & 0.06          & 0.00          & 0.00          & 0.00           & 0.00          & 0.00          & 0.00          & 0.00          & 0.00          & 0.00          & 0.00          & 0.00          \\ \hline
Eclipta                & 0.00          & 0.00          & \textbf{0.97} & 0.00          & 0.00          & 0.00          & 0.00           & 0.00          & 0.00          & 0.00          & 0.00          & 0.00          & 0.00          & 0.00          & 0.03          \\ \hline
Goosegrass             & 0.00          & 0.00          & 0.00          & \textbf{1.00} & 0.00          & 0.00          & 0.00           & 0.00          & 0.00          & 0.00          & 0.00          & 0.00          & 0.00          & 0.00          & 0.00          \\ \hline
Morningglory           & 0.00          & 0.00          & 0.00          & 0.00          & \textbf{1.00} & 0.00          & 0.00           & 0.00          & 0.00          & 0.00          & 0.00          & 0.00          & 0.00          & 0.00          & 0.00          \\ \hline
Nutsedge               & 0.00          & 0.00          & 0.00          & 0.00          & 0.00          & \textbf{1.00} & 0.00           & 0.00          & 0.00          & 0.00          & 0.00          & 0.00          & 0.00          & 0.00          & 0.00          \\
\hline

\end{tabular}}
\caption{The confusion matrix achieved by the googlenet model on the test subset of the CottonWeeds dataset.}
\label{tab:googlenet}
\end{table*}

\begin{table*}[!th]
\centering
\renewcommand{\arraystretch}{1.5}
\resizebox{1.0\textwidth}{!}{
\begin{tabular}{c|c|c|c|c|c|c|c|c|c|c|c|c|c|c|c}
\hline
\multicolumn{1}{l|}{} & Carpetweeds   & Crabgrass     & Eclipta       & Goosegrass    & Morningglory  & Nutsedge      & PalmerAmaranth & PricklySida   & Purslane      & Ragweed       & Sicklepod     & SpottedSpurge & SpurredAnoda  & Swinecress    & Waterhemp     \\ \hline
Carpetweeds            & \textbf{1.00} & 0.00          & 0.00          & 0.00          & 0.00          & 0.00          & 0.00           & 0.00          & 0.00          & 0.00          & 0.00          & 0.00          & 0.00          & 0.00          & 0.00          \\ \hline
Crabgrass              & 0.00          & \textbf{0.94} & 0.00          & 0.06          & 0.00          & 0.00          & 0.00           & 0.00          & 0.00          & 0.00          & 0.00          & 0.00          & 0.00          & 0.00          & 0.00          \\ \hline
Eclipta                & 0.03          & 0.00          & \textbf{0.97} & 0.00          & 0.00          & 0.00          & 0.00           & 0.00          & 0.00          & 0.00          & 0.00          & 0.00          & 0.00          & 0.00          & 0.00          \\ \hline
Goosegrass             & 0.00          & 0.00          & 0.00          & \textbf{1.00} & 0.00          & 0.00          & 0.00           & 0.00          & 0.00          & 0.00          & 0.00          & 0.00          & 0.00          & 0.00          & 0.00          \\ \hline
Morningglory           & 0.00          & 0.00          & 0.00          & 0.00          & \textbf{1.00} & 0.00          & 0.00           & 0.00          & 0.00          & 0.00          & 0.00          & 0.00          & 0.00          & 0.00          & 0.00          \\ \hline
Nutsedge               & 0.00          & 0.00          & 0.00          & 0.00          & 0.00          & \textbf{1.00} & 0.00           & 0.00          & 0.00          & 0.00          & 0.00          & 0.00          & 0.00          & 0.00          & 0.00          \\
\hline

\end{tabular}}
\caption{The confusion matrix achieved by the Xception model on the test subset of the CottonWeeds dataset.}
\label{tab:Xception}
\end{table*}

\begin{table*}[!th]
\centering
\renewcommand{\arraystretch}{1.5}
\resizebox{1.0\textwidth}{!}{
\begin{tabular}{c|c|c|c|c|c|c|c|c|c|c|c|c|c|c|c}
\hline
\multicolumn{1}{l|}{} & Carpetweeds   & Crabgrass     & Eclipta       & Goosegrass    & Morningglory  & Nutsedge      & PalmerAmaranth & PricklySida   & Purslane      & Ragweed       & Sicklepod     & SpottedSpurge & SpurredAnoda  & Swinecress    & Waterhemp     \\ \hline
Carpetweeds            & \textbf{1.00} & 0.00          & 0.00          & 0.00          & 0.00          & 0.00          & 0.00           & 0.00          & 0.00          & 0.00          & 0.00          & 0.00          & 0.00          & 0.00          & 0.00          \\ \hline
Crabgrass              & 0.00          & \textbf{1.00} & 0.00          & 0.00          & 0.00          & 0.00          & 0.00           & 0.00          & 0.00          & 0.00          & 0.00          & 0.00          & 0.00          & 0.00          & 0.00          \\ \hline
Eclipta                & 0.00          & 0.00          & \textbf{1.00} & 0.00          & 0.00          & 0.00          & 0.00           & 0.00          & 0.00          & 0.00          & 0.00          & 0.00          & 0.00          & 0.00          & 0.00          \\ \hline
Goosegrass             & 0.00          & 0.06          & 0.00          & \textbf{0.94} & 0.00          & 0.00          & 0.00           & 0.00          & 0.00          & 0.00          & 0.00          & 0.00          & 0.00          & 0.00          & 0.00          \\ \hline
Morningglory           & 0.00          & 0.00          & 0.00          & 0.00          & \textbf{1.00} & 0.00          & 0.00           & 0.00          & 0.00          & 0.00          & 0.00          & 0.00          & 0.00          & 0.00          & 0.00          \\ \hline
Nutsedge               & 0.00          & 0.00          & 0.00          & 0.00          & 0.00          & \textbf{1.00} & 0.00           & 0.00          & 0.00          & 0.00          & 0.00          & 0.00          & 0.00          & 0.00          & 0.00          \\
\hline

\end{tabular}}
\caption{The confusion matrix achieved by the dpn68	model on the test subset of the CottonWeeds dataset.}
\label{tab:dpn68}
\end{table*}

\begin{table*}[!th]
\centering
\renewcommand{\arraystretch}{1.5}
\resizebox{1.0\textwidth}{!}{
\begin{tabular}{c|c|c|c|c|c|c|c|c|c|c|c|c|c|c|c}
\hline
\multicolumn{1}{l|}{} & Carpetweeds   & Crabgrass     & Eclipta       & Goosegrass    & Morningglory  & Nutsedge      & PalmerAmaranth & PricklySida   & Purslane      & Ragweed       & Sicklepod     & SpottedSpurge & SpurredAnoda  & Swinecress    & Waterhemp     \\ \hline
Carpetweeds            & \textbf{0.99} & 0.00          & 0.01          & 0.00          & 0.00          & 0.00          & 0.00           & 0.00          & 0.00          & 0.00          & 0.00          & 0.00          & 0.00          & 0.00          & 0.00          \\ \hline
Crabgrass              & 0.00          & \textbf{0.88} & 0.00          & 0.12          & 0.00          & 0.00          & 0.00           & 0.00          & 0.00          & 0.00          & 0.00          & 0.00          & 0.00          & 0.00          & 0.00          \\ \hline
Eclipta                & 0.00          & 0.00          & \textbf{0.95} & 0.00          & 0.03          & 0.00          & 0.00           & 0.00          & 0.00          & 0.03          & 0.00          & 0.00          & 0.00          & 0.00          & 0.00          \\ \hline
Goosegrass             & 0.00          & 0.00          & 0.00          & \textbf{0.94} & 0.00          & 0.06          & 0.00           & 0.00          & 0.00          & 0.00          & 0.00          & 0.00          & 0.00          & 0.00          & 0.00          \\ \hline
Morningglory           & 0.00          & 0.00          & 0.00          & 0.00          & \textbf{0.99} & 0.00          & 0.01           & 0.00          & 0.00          & 0.00          & 0.01          & 0.00          & 0.00          & 0.00          & 0.00          \\ \hline
Nutsedge               & 0.00          & 0.00          & 0.00          & 0.00          & 0.00          & \textbf{1.00} & 0.00           & 0.00          & 0.00          & 0.00          & 0.00          & 0.00          & 0.00          & 0.00          & 0.00          \\
\hline

\end{tabular}}
\caption{The confusion matrix achieved by the MnasNet model on the test subset of the CottonWeeds dataset.}
\label{tab:mnasnet}
\end{table*}

\begin{table*}[!th]
\centering
\renewcommand{\arraystretch}{1.5}
\resizebox{1.0\textwidth}{!}{
\begin{tabular}{c|c|c|c|c|c|c|c|c|c|c|c|c|c|c|c}
\hline
\multicolumn{1}{l|}{} & Carpetweeds   & Crabgrass     & Eclipta       & Goosegrass    & Morningglory  & Nutsedge      & PalmerAmaranth & PricklySida   & Purslane      & Ragweed       & Sicklepod     & SpottedSpurge & SpurredAnoda  & Swinecress    & Waterhemp     \\ \hline
Carpetweeds            & \textbf{1.00} & 0.00          & 0.00          & 0.00          & 0.00          & 0.00          & 0.00           & 0.00          & 0.00          & 0.00          & 0.00          & 0.00          & 0.00          & 0.00          & 0.00          \\ \hline
Crabgrass              & 0.00          & \textbf{0.94} & 0.00          & 0.06          & 0.00          & 0.00          & 0.00           & 0.00          & 0.00          & 0.00          & 0.00          & 0.00          & 0.00          & 0.00          & 0.00          \\ \hline
Eclipta                & 0.00          & 0.00          & \textbf{1.00} & 0.00          & 0.00          & 0.00          & 0.00           & 0.00          & 0.00          & 0.00          & 0.00          & 0.00          & 0.00          & 0.00          & 0.00          \\ \hline
Goosegrass             & 0.00          & 0.03          & 0.00          & \textbf{0.97} & 0.00          & 0.00          & 0.00           & 0.00          & 0.00          & 0.00          & 0.00          & 0.00          & 0.00          & 0.00          & 0.00          \\ \hline
Morningglory           & 0.00          & 0.00          & 0.00          & 0.00          & \textbf{1.00} & 0.00          & 0.00           & 0.00          & 0.00          & 0.00          & 0.00          & 0.00          & 0.00          & 0.00          & 0.00          \\ \hline
Nutsedge               & 0.00          & 0.00          & 0.00          & 0.00          & 0.00          & \textbf{1.00} & 0.00           & 0.00          & 0.00          & 0.00          & 0.00          & 0.00          & 0.00          & 0.00          & 0.00          \\
\hline

\end{tabular}}
\caption{The confusion matrix achieved by the ResNet18 model on the test subset of the CottonWeeds dataset.}
\label{tab:ResNet18}
\end{table*}

\begin{table*}[!th]
\centering
\renewcommand{\arraystretch}{1.5}
\resizebox{1.0\textwidth}{!}{
\begin{tabular}{c|c|c|c|c|c|c|c|c|c|c|c|c|c|c|c}
\hline
\multicolumn{1}{l|}{} & Carpetweeds   & Crabgrass     & Eclipta       & Goosegrass    & Morningglory  & Nutsedge      & PalmerAmaranth & PricklySida   & Purslane      & Ragweed       & Sicklepod     & SpottedSpurge & SpurredAnoda  & Swinecress    & Waterhemp     \\ \hline
Carpetweeds            & \textbf{1.00} & 0.00          & 0.00          & 0.00          & 0.00          & 0.00          & 0.00           & 0.00          & 0.00          & 0.00          & 0.00          & 0.00          & 0.00          & 0.00          & 0.00          \\ \hline
Crabgrass              & 0.00          & \textbf{1.00} & 0.00          & 0.00          & 0.00          & 0.00          & 0.00           & 0.00          & 0.00          & 0.00          & 0.00          & 0.00          & 0.00          & 0.00          & 0.00          \\ \hline
Eclipta                & 0.00          & 0.00          & \textbf{1.00} & 0.00          & 0.00          & 0.00          & 0.00           & 0.00          & 0.00          & 0.00          & 0.00          & 0.00          & 0.00          & 0.00          & 0.00          \\ \hline
Goosegrass             & 0.00          & 0.03          & 0.00          & \textbf{0.97} & 0.00          & 0.00          & 0.00           & 0.00          & 0.00          & 0.00          & 0.00          & 0.00          & 0.00          & 0.00          & 0.00          \\ \hline
Morningglory           & 0.00          & 0.00          & 0.00          & 0.00          & \textbf{1.00} & 0.00          & 0.00           & 0.00          & 0.00          & 0.00          & 0.00          & 0.00          & 0.00          & 0.00          & 0.00          \\ \hline
Nutsedge               & 0.00          & 0.00          & 0.00          & 0.00          & 0.00          & \textbf{1.00} & 0.00           & 0.00          & 0.00          & 0.00          & 0.00          & 0.00          & 0.00          & 0.00          & 0.00          \\
\hline

\end{tabular}}
\caption{The confusion matrix achieved by the ResNet50 model on the test subset of the CottonWeeds dataset.}
\label{tab:ResNet50}
\end{table*}

\begin{table*}[!th]
\renewcommand{\arraystretch}{1.5}
\resizebox{1.0\textwidth}{!}{
\begin{tabular}{c|c|c|c|c|c|c|c|c|c|c|c|c|c|c|c}
\hline
               & Carpetweeds   & Crabgrass     & Eclipta       & Goosegrass    & Morningglory  & Nutsedge      & PalmerAmaranth & PricklySida   & Purslane      & Ragweed       & Sicklepod     & SpottedSpurge & SpurredAnoda  & Swinecress    & Waterhemp     \\ \hline
Carpetweeds    & \textbf{1.00} & 0.00          & 0.00          & 0.00          & 0.00          & 0.00          & 0.00           & 0.00          & 0.00          & 0.00          & 0.00          & 0.00          & 0.00          & 0.00          & 0.00          \\ \hline
Crabgrass      & 0.00          & \textbf{1.00} & 0.00          & 0.00          & 0.00          & 0.00          & 0.00           & 0.00          & 0.00          & 0.00          & 0.00          & 0.00          & 0.00          & 0.00          & 0.00          \\ \hline
Eclipta        & 0.00          & 0.00          & \textbf{1.00} & 0.00          & 0.00          & 0.00          & 0.00           & 0.00          & 0.00          & 0.00          & 0.00          & 0.00          & 0.00          & 0.00          & 0.00          \\ \hline
Goosegrass     & 0.00          & 0.03          & 0.00          & \textbf{0.97} & 0.00          & 0.00          & 0.00           & 0.00          & 0.00          & 0.00          & 0.00          & 0.00          & 0.00          & 0.00          & 0.00          \\ \hline
Morningglory   & 0.00          & 0.00          & 0.00          & 0.00          & \textbf{1.00} & 0.00          & 0.00           & 0.00          & 0.00          & 0.00          & 0.00          & 0.00          & 0.00          & 0.00          & 0.00          \\ \hline
Nutsedge       & 0.00          & 0.00          & 0.00          & 0.00          & 0.00          & \textbf{1.00} & 0.00           & 0.00          & 0.00          & 0.00          & 0.00          & 0.00          & 0.00          & 0.00          & 0.00          \\
\hline

\end{tabular}}
\caption{The confusion matrix achieved by the ResNet-101 model on the test subset of the CottonWeeds dataset.}
\label{tab:resnet_cm}
\end{table*}

\begin{table*}[!th]
\centering
\renewcommand{\arraystretch}{1.5}
\resizebox{1.0\textwidth}{!}{
\begin{tabular}{c|c|c|c|c|c|c|c|c|c|c|c|c|c|c|c}
\hline
\multicolumn{1}{l|}{} & Carpetweeds   & Crabgrass     & Eclipta       & Goosegrass    & Morningglory  & Nutsedge      & PalmerAmaranth & PricklySida   & Purslane      & Ragweed       & Sicklepod     & SpottedSpurge & SpurredAnoda  & Swinecress    & Waterhemp     \\ \hline
Carpetweeds            & \textbf{1.00} & 0.00          & 0.00          & 0.00          & 0.00          & 0.00          & 0.00           & 0.00          & 0.00          & 0.00          & 0.00          & 0.00          & 0.00          & 0.00          & 0.00          \\ \hline
Crabgrass              & 0.00          & \textbf{1.00} & 0.00          & 0.00          & 0.00          & 0.00          & 0.00           & 0.00          & 0.00          & 0.00          & 0.00          & 0.00          & 0.00          & 0.00          & 0.00          \\ \hline
Eclipta                & 0.05          & 0.00          & \textbf{0.95} & 0.00          & 0.00          & 0.00          & 0.00           & 0.00          & 0.00          & 0.00          & 0.00          & 0.00          & 0.00          & 0.00          & 0.00          \\ \hline
Goosegrass             & 0.00          & 0.00          & 0.00          & \textbf{0.97} & 0.00          & 0.00          & 0.00           & 0.00          & 0.00          & 0.00          & 0.00          & 0.00          & 0.00          & 0.00          & 0.03          \\ \hline
Morningglory           & 0.00          & 0.00          & 0.00          & 0.00          & \textbf{0.99} & 0.00          & 0.00           & 0.00          & 0.01          & 0.00          & 0.00          & 0.00          & 0.00          & 0.00          & 0.00          \\ \hline
Nutsedge               & 0.00          & 0.00          & 0.00          & 0.00          & 0.00          & \textbf{1.00} & 0.00           & 0.00          & 0.00          & 0.00          & 0.00          & 0.00          & 0.00          & 0.00          & 0.00          \\
\hline

\end{tabular}}
\caption{The confusion matrix achieved by the VGG11 model on the test subset of the CottonWeeds dataset.}
\label{tab:VGG11}
\end{table*}

\begin{table*}[!th]
\centering
\renewcommand{\arraystretch}{1.5}
\resizebox{1.0\textwidth}{!}{
\begin{tabular}{c|c|c|c|c|c|c|c|c|c|c|c|c|c|c|c}
\hline
\multicolumn{1}{l|}{} & Carpetweeds   & Crabgrass     & Eclipta       & Goosegrass    & Morningglory  & Nutsedge      & PalmerAmaranth & PricklySida   & Purslane      & Ragweed       & Sicklepod     & SpottedSpurge & SpurredAnoda  & Swinecress    & Waterhemp     \\ \hline
Carpetweeds            & \textbf{1.00} & 0.00          & 0.00          & 0.00          & 0.00          & 0.00          & 0.00           & 0.00          & 0.00          & 0.00          & 0.00          & 0.00          & 0.00          & 0.00          & 0.00          \\ \hline
Crabgrass              & 0.00          & \textbf{1.00} & 0.00          & 0.00          & 0.00          & 0.00          & 0.00           & 0.00          & 0.00          & 0.00          & 0.00          & 0.00          & 0.00          & 0.00          & 0.00          \\ \hline
Eclipta                & 0.05          & 0.00          & \textbf{0.92} & 0.00          & 0.00          & 0.00          & 0.00           & 0.00          & 0.00          & 0.00          & 0.00          & 0.00          & 0.00          & 0.00          & 0.03          \\ \hline
Goosegrass             & 0.00          & 0.03          & 0.00          & \textbf{0.97} & 0.00          & 0.00          & 0.00           & 0.00          & 0.00          & 0.00          & 0.00          & 0.00          & 0.00          & 0.00          & 0.00          \\ \hline
Morningglory           & 0.00          & 0.00          & 0.00          & 0.00          & \textbf{1.00} & 0.00          & 0.00           & 0.00          & 0.00          & 0.00          & 0.00          & 0.00          & 0.00          & 0.00          & 0.00          \\ \hline
Nutsedge               & 0.00          & 0.00          & 0.00          & 0.00          & 0.00          & \textbf{1.00} & 0.00           & 0.00          & 0.00          & 0.00          & 0.00          & 0.00          & 0.00          & 0.00          & 0.00          \\
\hline

\end{tabular}}
\caption{The confusion matrix achieved by the VGG16 model on the test subset of the CottonWeeds dataset.}
\label{tab:VGG16}
\end{table*}

\begin{table*}[!th]
\centering
\renewcommand{\arraystretch}{1.5}
\resizebox{1.0\textwidth}{!}{
\begin{tabular}{c|c|c|c|c|c|c|c|c|c|c|c|c|c|c|c}
\hline
\multicolumn{1}{l|}{} & Carpetweeds   & Crabgrass     & Eclipta       & Goosegrass    & Morningglory  & Nutsedge      & PalmerAmaranth & PricklySida   & Purslane      & Ragweed       & Sicklepod     & SpottedSpurge & SpurredAnoda  & Swinecress    & Waterhemp     \\ \hline
Carpetweeds            & \textbf{0.99} & 0.00          & 0.00          & 0.00          & 0.00          & 0.00          & 0.00           & 0.00          & 0.01          & 0.00          & 0.00          & 0.00          & 0.00          & 0.00          & 0.00          \\ \hline
Crabgrass              & 0.00          & \textbf{1.00} & 0.00          & 0.00          & 0.00          & 0.00          & 0.00           & 0.00          & 0.00          & 0.00          & 0.00          & 0.00          & 0.00          & 0.00          & 0.00          \\ \hline
Eclipta                & 0.00          & 0.00          & \textbf{0.92} & 0.03          & 0.00          & 0.00          & 0.00           & 0.00          & 0.00          & 0.00          & 0.00          & 0.00          & 0.00          & 0.00          & 0.05          \\ \hline
Goosegrass             & 0.00          & 0.03          & 0.00          & \textbf{0.97} & 0.00          & 0.00          & 0.00           & 0.00          & 0.00          & 0.00          & 0.00          & 0.00          & 0.00          & 0.00          & 0.00          \\ \hline
Morningglory           & 0.00          & 0.00          & 0.00          & 0.00          & \textbf{1.00} & 0.00          & 0.00           & 0.00          & 0.00          & 0.00          & 0.00          & 0.00          & 0.00          & 0.00          & 0.00          \\ \hline
Nutsedge               & 0.00          & 0.00          & 0.00          & 0.00          & 0.00          & \textbf{1.00} & 0.00           & 0.00          & 0.00          & 0.00          & 0.00          & 0.00          & 0.00          & 0.00          & 0.00          \\
\hline

\end{tabular}}
\caption{The confusion matrix achieved by the VGG19 model on the test subset of the CottonWeeds dataset.}
\label{tab:VGG19}
\end{table*}

\begin{table*}[!th]
\centering
\renewcommand{\arraystretch}{1.5}
\resizebox{1.0\textwidth}{!}{
\begin{tabular}{c|c|c|c|c|c|c|c|c|c|c|c|c|c|c|c}
\hline
\multicolumn{1}{l|}{} & Carpetweeds   & Crabgrass     & Eclipta       & Goosegrass    & Morningglory  & Nutsedge      & PalmerAmaranth & PricklySida   & Purslane      & Ragweed       & Sicklepod     & SpottedSpurge & SpurredAnoda  & Swinecress    & Waterhemp     \\ \hline
Carpetweeds            & \textbf{0.97} & 0.00          & 0.01          & 0.00          & 0.02          & 0.00          & 0.00           & 0.00          & 0.00          & 0.00          & 0.00          & 0.00          & 0.00          & 0.00          & 0.00          \\ \hline
Crabgrass              & 0.00          & \textbf{0.94} & 0.00          & 0.06          & 0.00          & 0.00          & 0.00           & 0.00          & 0.00          & 0.00          & 0.00          & 0.00          & 0.00          & 0.00          & 0.00          \\ \hline
Eclipta                & 0.08          & 0.00          & \textbf{0.85} & 0.00          & 0.00          & 0.00          & 0.00           & 0.03          & 0.00          & 0.00          & 0.00          & 0.00          & 0.03          & 0.00          & 0.03          \\ \hline
Goosegrass             & 0.00          & 0.03          & 0.00          & \textbf{0.97} & 0.00          & 0.00          & 0.00           & 0.00          & 0.00          & 0.00          & 0.00          & 0.00          & 0.00          & 0.00          & 0.00          \\ \hline
Morningglory           & 0.00          & 0.00          & 0.00          & 0.00          & \textbf{0.99} & 0.00          & 0.00           & 0.00          & 0.01          & 0.00          & 0.00          & 0.00          & 0.00          & 0.00          & 0.00          \\ \hline
Nutsedge               & 0.00          & 0.00          & 0.00          & 0.00          & 0.00          & \textbf{1.00} & 0.00           & 0.00          & 0.00          & 0.00          & 0.00          & 0.00          & 0.00          & 0.00          & 0.00          \\
\hline

\end{tabular}}
\caption{The confusion matrix achieved by the Densenet121 model on the test subset of the CottonWeeds dataset.}
\label{tab:Densenet121}
\end{table*}

\begin{table*}[!th]
\centering
\renewcommand{\arraystretch}{1.5}
\resizebox{1.0\textwidth}{!}{
\begin{tabular}{c|c|c|c|c|c|c|c|c|c|c|c|c|c|c|c}
\hline
\multicolumn{1}{l|}{} & Carpetweeds   & Crabgrass     & Eclipta       & Goosegrass    & Morningglory  & Nutsedge      & PalmerAmaranth & PricklySida   & Purslane      & Ragweed       & Sicklepod     & SpottedSpurge & SpurredAnoda  & Swinecress    & Waterhemp     \\ \hline
Carpetweeds            & \textbf{0.97} & 0.00          & 0.01          & 0.00          & 0.02          & 0.00          & 0.00           & 0.00          & 0.00          & 0.00          & 0.00          & 0.00          & 0.00          & 0.00          & 0.00          \\ \hline
Crabgrass              & 0.00          & \textbf{0.94} & 0.00          & 0.06          & 0.00          & 0.00          & 0.00           & 0.00          & 0.00          & 0.00          & 0.00          & 0.00          & 0.00          & 0.00          & 0.00          \\ \hline
Eclipta                & 0.08          & 0.00          & \textbf{0.85} & 0.00          & 0.00          & 0.00          & 0.00           & 0.03          & 0.00          & 0.00          & 0.00          & 0.00          & 0.03          & 0.00          & 0.03          \\ \hline
Goosegrass             & 0.00          & 0.03          & 0.00          & \textbf{0.97} & 0.00          & 0.00          & 0.00           & 0.00          & 0.00          & 0.00          & 0.00          & 0.00          & 0.00          & 0.00          & 0.00          \\ \hline
Morningglory           & 0.00          & 0.00          & 0.00          & 0.00          & \textbf{0.99} & 0.00          & 0.00           & 0.00          & 0.01          & 0.00          & 0.00          & 0.00          & 0.00          & 0.00          & 0.00          \\ \hline
Nutsedge               & 0.00          & 0.00          & 0.00          & 0.00          & 0.00          & \textbf{1.00} & 0.00           & 0.00          & 0.00          & 0.00          & 0.00          & 0.00          & 0.00          & 0.00          & 0.00          \\
\hline

\end{tabular}}
\caption{The confusion matrix achieved by the Densenet161 model on the test subset of the CottonWeeds dataset.}
\label{tab:Densenet161}
\end{table*}

\begin{table*}[!th]
\centering
\renewcommand{\arraystretch}{1.5}
\resizebox{1.0\textwidth}{!}{
\begin{tabular}{c|c|c|c|c|c|c|c|c|c|c|c|c|c|c|c}
\hline
\multicolumn{1}{l|}{} & Carpetweeds   & Crabgrass     & Eclipta       & Goosegrass    & Morningglory  & Nutsedge      & PalmerAmaranth & PricklySida   & Purslane      & Ragweed       & Sicklepod     & SpottedSpurge & SpurredAnoda  & Swinecress    & Waterhemp     \\ \hline
Carpetweeds            & \textbf{0.97} & 0.00          & 0.01          & 0.00          & 0.02          & 0.00          & 0.00           & 0.00          & 0.00          & 0.00          & 0.00          & 0.00          & 0.00          & 0.00          & 0.00          \\ \hline
Crabgrass              & 0.00          & \textbf{0.94} & 0.00          & 0.06          & 0.00          & 0.00          & 0.00           & 0.00          & 0.00          & 0.00          & 0.00          & 0.00          & 0.00          & 0.00          & 0.00          \\ \hline
Eclipta                & 0.08          & 0.00          & \textbf{0.85} & 0.00          & 0.00          & 0.00          & 0.00           & 0.03          & 0.00          & 0.00          & 0.00          & 0.00          & 0.03          & 0.00          & 0.03          \\ \hline
Goosegrass             & 0.00          & 0.03          & 0.00          & \textbf{0.97} & 0.00          & 0.00          & 0.00           & 0.00          & 0.00          & 0.00          & 0.00          & 0.00          & 0.00          & 0.00          & 0.00          \\ \hline
Morningglory           & 0.00          & 0.00          & 0.00          & 0.00          & \textbf{0.99} & 0.00          & 0.00           & 0.00          & 0.01          & 0.00          & 0.00          & 0.00          & 0.00          & 0.00          & 0.00          \\ \hline
Nutsedge               & 0.00          & 0.00          & 0.00          & 0.00          & 0.00          & \textbf{1.00} & 0.00           & 0.00          & 0.00          & 0.00          & 0.00          & 0.00          & 0.00          & 0.00          & 0.00          \\
\hline

\end{tabular}}
\caption{The confusion matrix achieved by the Densenet161 model on the test subset of the CottonWeeds dataset.}
\label{tab:Densenet161}
\end{table*}

\begin{table*}[!th]
\centering
\renewcommand{\arraystretch}{1.5}
\resizebox{1.0\textwidth}{!}{
\begin{tabular}{c|c|c|c|c|c|c|c|c|c|c|c|c|c|c|c}
\hline
\multicolumn{1}{l|}{} & Carpetweeds   & Crabgrass     & Eclipta       & Goosegrass    & Morningglory  & Nutsedge      & PalmerAmaranth & PricklySida   & Purslane      & Ragweed       & Sicklepod     & SpottedSpurge & SpurredAnoda  & Swinecress    & Waterhemp     \\ \hline
Carpetweeds            & \textbf{1.00} & 0.00          & 0.00          & 0.00          & 0.00          & 0.00          & 0.00           & 0.00          & 0.00          & 0.00          & 0.00          & 0.00          & 0.00          & 0.00          & 0.00          \\ \hline
Crabgrass              & 0.00          & \textbf{1.00} & 0.00          & 0.00          & 0.00          & 0.00          & 0.00           & 0.00          & 0.00          & 0.00          & 0.00          & 0.00          & 0.00          & 0.00          & 0.00          \\ \hline
Eclipta                & 0.00          & 0.00          & \textbf{1.00} & 0.00          & 0.00          & 0.00          & 0.00           & 0.00          & 0.00          & 0.00          & 0.00          & 0.00          & 0.00          & 0.00          & 0.00          \\ \hline
Goosegrass             & 0.00          & 0.00          & 0.00          & \textbf{1.00} & 0.00          & 0.00          & 0.00           & 0.00          & 0.00          & 0.00          & 0.00          & 0.00          & 0.00          & 0.00          & 0.00          \\ \hline
Morningglory           & 0.00          & 0.00          & 0.00          & 0.00          & \textbf{1.00} & 0.00          & 0.00           & 0.00          & 0.00          & 0.00          & 0.00          & 0.00          & 0.00          & 0.00          & 0.00          \\ \hline
Nutsedge               & 0.00          & 0.00          & 0.00          & 0.00          & 0.00          & \textbf{1.00} & 0.00           & 0.00          & 0.00          & 0.00          & 0.00          & 0.00          & 0.00          & 0.00          & 0.00          \\
\hline

\end{tabular}}
\caption{The confusion matrix achieved by the Inception v3 model on the test subset of the CottonWeeds dataset.}
\label{tab:Inception-v3}
\end{table*}

\begin{table*}[!th]
\centering
\renewcommand{\arraystretch}{1.5}
\resizebox{1.0\textwidth}{!}{
\begin{tabular}{c|c|c|c|c|c|c|c|c|c|c|c|c|c|c|c}
\hline
\multicolumn{1}{l|}{} & Carpetweeds   & Crabgrass     & Eclipta       & Goosegrass    & Morningglory  & Nutsedge      & PalmerAmaranth & PricklySida   & Purslane      & Ragweed       & Sicklepod     & SpottedSpurge & SpurredAnoda  & Swinecress    & Waterhemp     \\ \hline
Carpetweeds            & \textbf{1.00} & 0.00          & 0.00          & 0.00          & 0.00          & 0.00          & 0.00           & 0.00          & 0.00          & 0.00          & 0.00          & 0.00          & 0.00          & 0.00          & 0.00          \\ \hline
Crabgrass              & 0.00          & \textbf{0.94} & 0.00          & 0.06          & 0.00          & 0.00          & 0.00           & 0.00          & 0.00          & 0.00          & 0.00          & 0.00          & 0.00          & 0.00          & 0.00          \\ \hline
Eclipta                & 0.00          & 0.03          & \textbf{0.95} & 0.00          & 0.00          & 0.00          & 0.00           & 0.00          & 0.00          & 0.03          & 0.00          & 0.00          & 0.00          & 0.00          & 0.00          \\ \hline
Goosegrass             & 0.00          & 0.06          & 0.00          & \textbf{0.94} & 0.00          & 0.00          & 0.00           & 0.00          & 0.00          & 0.00          & 0.00          & 0.00          & 0.00          & 0.00          & 0.00          \\ \hline
Morningglory           & 0.00          & 0.00          & 0.00          & 0.00          & \textbf{0.99} & 0.00          & 0.00           & 0.00          & 0.00          & 0.00          & 0.00          & 0.00          & 0.01          & 0.00          & 0.00          \\ \hline
Nutsedge               & 0.00          & 0.00          & 0.00          & 0.05          & 0.00          & \textbf{0.95} & 0.00           & 0.00          & 0.00          & 0.00          & 0.00          & 0.00          & 0.00          & 0.00          & 0.00          \\
\hline

\end{tabular}}
\caption{The confusion matrix achieved by the Inception v4 model on the test subset of the CottonWeeds dataset.}
\label{tab:Inception-v4}
\end{table*}

\begin{table*}[!th]
\centering
\renewcommand{\arraystretch}{1.5}
\resizebox{1.0\textwidth}{!}{
\begin{tabular}{c|c|c|c|c|c|c|c|c|c|c|c|c|c|c|c}
\hline
\multicolumn{1}{l|}{} & Carpetweeds   & Crabgrass     & Eclipta       & Goosegrass    & Morningglory  & Nutsedge      & PalmerAmaranth & PricklySida   & Purslane      & Ragweed       & Sicklepod     & SpottedSpurge & SpurredAnoda  & Swinecress    & Waterhemp     \\ \hline
Carpetweeds            & \textbf{0.99} & 0.00          & 0.00          & 0.00          & 0.00          & 0.00          & 0.00           & 0.00          & 0.01          & 0.00          & 0.00          & 0.00          & 0.00          & 0.00          & 0.00          \\ \hline
Crabgrass              & 0.00          & \textbf{0.94} & 0.00          & 0.06          & 0.00          & 0.00          & 0.00           & 0.00          & 0.00          & 0.00          & 0.00          & 0.00          & 0.00          & 0.00          & 0.00          \\ \hline
Eclipta                & 0.00          & 0.00          & \textbf{0.97} & 0.00          & 0.00          & 0.00          & 0.00           & 0.00          & 0.00          & 0.00          & 0.00          & 0.00          & 0.00          & 0.00          & 0.03          \\ \hline
Goosegrass             & 0.03          & 0.00          & 0.00          & \textbf{0.97} & 0.00          & 0.00          & 0.00           & 0.00          & 0.00          & 0.00          & 0.00          & 0.00          & 0.00          & 0.00          & 0.00          \\ \hline
Morningglory           & 0.00          & 0.00          & 0.00          & 0.00          & \textbf{1.00} & 0.00          & 0.00           & 0.00          & 0.00          & 0.00          & 0.00          & 0.00          & 0.00          & 0.00          & 0.00          \\ \hline
Nutsedge               & 0.00          & 0.00          & 0.00          & 0.00          & 0.00          & \textbf{1.00} & 0.00           & 0.00          & 0.00          & 0.00          & 0.00          & 0.00          & 0.00          & 0.00          & 0.00          \\
\hline

\end{tabular}}
\caption{The confusion matrix achieved by the Inception-ResNet v2 model on the test subset of the CottonWeeds dataset.}
\label{tab:Inception-ResNet-v2}
\end{table*}

\begin{table*}[!th]
\centering
\renewcommand{\arraystretch}{1.5}
\resizebox{1.0\textwidth}{!}{
\begin{tabular}{c|c|c|c|c|c|c|c|c|c|c|c|c|c|c|c}
\hline
\multicolumn{1}{l|}{} & Carpetweeds   & Crabgrass     & Eclipta       & Goosegrass    & Morningglory  & Nutsedge      & PalmerAmaranth & PricklySida   & Purslane      & Ragweed       & Sicklepod     & SpottedSpurge & SpurredAnoda  & Swinecress    & Waterhemp     \\ \hline
Carpetweeds            & \textbf{1.00} & 0.00          & 0.00          & 0.00          & 0.00          & 0.00          & 0.00           & 0.00          & 0.00          & 0.00          & 0.00          & 0.00          & 0.00          & 0.00          & 0.00          \\ \hline
Crabgrass              & 0.00          & \textbf{1.00} & 0.00          & 0.00          & 0.00          & 0.00          & 0.00           & 0.00          & 0.00          & 0.00          & 0.00          & 0.00          & 0.00          & 0.00          & 0.00          \\ \hline
Eclipta                & 0.00          & 0.00          & \textbf{1.00} & 0.00          & 0.00          & 0.00          & 0.00           & 0.00          & 0.00          & 0.00          & 0.00          & 0.00          & 0.00          & 0.00          & 0.00          \\ \hline
Goosegrass             & 0.00          & 0.06          & 0.00          & \textbf{0.94} & 0.00          & 0.00          & 0.00           & 0.00          & 0.00          & 0.00          & 0.00          & 0.00          & 0.00          & 0.00          & 0.00          \\ \hline
Morningglory           & 0.00          & 0.00          & 0.00          & 0.00          & \textbf{0.99} & 0.00          & 0.01           & 0.00          & 0.00          & 0.00          & 0.01          & 0.00          & 0.00          & 0.00          & 0.00          \\ \hline
Nutsedge               & 0.00          & 0.00          & 0.00          & 0.00          & 0.00          & \textbf{1.00} & 0.00           & 0.00          & 0.00          & 0.00          & 0.00          & 0.00          & 0.00          & 0.00          & 0.00          \\
\hline

\end{tabular}}
\caption{The confusion matrix achieved by the MobilenetV2 model on the test subset of the CottonWeeds dataset.}
\label{tab:MobilenetV2}
\end{table*}

\begin{table*}[!th]
\centering
\renewcommand{\arraystretch}{1.5}
\resizebox{1.0\textwidth}{!}{
\begin{tabular}{c|c|c|c|c|c|c|c|c|c|c|c|c|c|c|c}
\hline
\multicolumn{1}{l|}{} & Carpetweeds   & Crabgrass     & Eclipta       & Goosegrass    & Morningglory  & Nutsedge      & PalmerAmaranth & PricklySida   & Purslane      & Ragweed       & Sicklepod     & SpottedSpurge & SpurredAnoda  & Swinecress    & Waterhemp     \\ \hline
Carpetweeds            & \textbf{1.00} & 0.00          & 0.00          & 0.00          & 0.00          & 0.00          & 0.00           & 0.00          & 0.00          & 0.00          & 0.00          & 0.00          & 0.00          & 0.00          & 0.00          \\ \hline
Crabgrass              & 0.00          & \textbf{0.94} & 0.00          & 0.06          & 0.00          & 0.00          & 0.00           & 0.00          & 0.00          & 0.00          & 0.00          & 0.00          & 0.00          & 0.00          & 0.00          \\ \hline
Eclipta                & 0.05          & 0.00          & \textbf{0.92} & 0.00          & 0.00          & 0.00          & 0.00           & 0.00          & 0.00          & 0.00          & 0.00          & 0.00          & 0.00          & 0.00          & 0.03          \\ \hline
Goosegrass             & 0.00          & 0.00          & 0.00          & \textbf{1.00} & 0.00          & 0.00          & 0.00           & 0.00          & 0.00          & 0.00          & 0.00          & 0.00          & 0.00          & 0.00          & 0.00          \\ \hline
Morningglory           & 0.00          & 0.00          & 0.00          & 0.00          & \textbf{0.99} & 0.00          & 0.00           & 0.00          & 0.00          & 0.00          & 0.01          & 0.00          & 0.00          & 0.00          & 0.00          \\ \hline
Nutsedge               & 0.00          & 0.00          & 0.00          & 0.00          & 0.00          & \textbf{1.00} & 0.00           & 0.00          & 0.00          & 0.00          & 0.00          & 0.00          & 0.00          & 0.00          & 0.00          \\
\hline

\end{tabular}}
\caption{The confusion matrix achieved by the MobilenetV3-small model on the test subset of the CottonWeeds dataset.}
\label{tab:MobilenetV3-small}
\end{table*}

\begin{table*}[!th]
\centering
\renewcommand{\arraystretch}{1.5}
\resizebox{1.0\textwidth}{!}{
\begin{tabular}{c|c|c|c|c|c|c|c|c|c|c|c|c|c|c|c}
\hline
\multicolumn{1}{l|}{} & Carpetweeds   & Crabgrass     & Eclipta       & Goosegrass    & Morningglory  & Nutsedge      & PalmerAmaranth & PricklySida   & Purslane      & Ragweed       & Sicklepod     & SpottedSpurge & SpurredAnoda  & Swinecress    & Waterhemp     \\ \hline
Carpetweeds            & \textbf{1.00} & 0.00          & 0.00          & 0.00          & 0.00          & 0.00          & 0.00           & 0.00          & 0.00          & 0.00          & 0.00          & 0.00          & 0.00          & 0.00          & 0.00          \\ \hline
Crabgrass              & 0.00          & \textbf{0.94} & 0.00          & 0.06          & 0.00          & 0.00          & 0.00           & 0.00          & 0.00          & 0.00          & 0.00          & 0.00          & 0.00          & 0.00          & 0.00          \\ \hline
Eclipta                & 0.00          & 0.00          & \textbf{1.00} & 0.00          & 0.00          & 0.00          & 0.00           & 0.00          & 0.00          & 0.00          & 0.00          & 0.00          & 0.00          & 0.00          & 0.00          \\ \hline
Goosegrass             & 0.00          & 0.00          & 0.00          & \textbf{1.00} & 0.00          & 0.00          & 0.00           & 0.00          & 0.00          & 0.00          & 0.00          & 0.00          & 0.00          & 0.00          & 0.00          \\ \hline
Morningglory           & 0.00          & 0.00          & 0.00          & 0.00          & \textbf{1.00} & 0.00          & 0.00           & 0.00          & 0.00          & 0.00          & 0.00          & 0.00          & 0.00          & 0.00          & 0.00          \\ \hline
Nutsedge               & 0.00          & 0.00          & 0.00          & 0.00          & 0.00          & \textbf{1.00} & 0.00           & 0.00          & 0.00          & 0.00          & 0.00          & 0.00          & 0.00          & 0.00          & 0.00          \\
\hline

\end{tabular}}
\caption{The confusion matrix achieved by the MobilenetV3-large model on the test subset of the CottonWeeds dataset.}
\label{tab:MobilenetV3-large}
\end{table*}

\begin{table*}[!th]
\centering
\renewcommand{\arraystretch}{1.5}
\resizebox{1.0\textwidth}{!}{
\begin{tabular}{c|c|c|c|c|c|c|c|c|c|c|c|c|c|c|c}
\hline
\multicolumn{1}{l|}{} & Carpetweeds   & Crabgrass     & Eclipta       & Goosegrass    & Morningglory  & Nutsedge      & PalmerAmaranth & PricklySida   & Purslane      & Ragweed       & Sicklepod     & SpottedSpurge & SpurredAnoda  & Swinecress    & Waterhemp     \\ \hline
Carpetweeds            & \textbf{0.99} & 0.00          & 0.01          & 0.00          & 0.00          & 0.00          & 0.00           & 0.00          & 0.00          & 0.00          & 0.00          & 0.00          & 0.00          & 0.00          & 0.00          \\ \hline
Crabgrass              & 0.00          & \textbf{0.94} & 0.00          & 0.06          & 0.00          & 0.00          & 0.00           & 0.00          & 0.00          & 0.00          & 0.00          & 0.00          & 0.00          & 0.00          & 0.00          \\ \hline
Eclipta                & 0.00          & 0.00          & \textbf{0.95} & 0.03          & 0.00          & 0.00          & 0.00           & 0.00          & 0.00          & 0.00          & 0.00          & 0.00          & 0.00          & 0.00          & 0.03          \\ \hline
Goosegrass             & 0.00          & 0.00          & 0.00          & \textbf{1.00} & 0.00          & 0.00          & 0.00           & 0.00          & 0.00          & 0.00          & 0.00          & 0.00          & 0.00          & 0.00          & 0.00          \\ \hline
Morningglory           & 0.00          & 0.00          & 0.00          & 0.00          & \textbf{0.98} & 0.00          & 0.01           & 0.00          & 0.01          & 0.00          & 0.01          & 0.00          & 0.00          & 0.00          & 0.00          \\ \hline
Nutsedge               & 0.00          & 0.00          & 0.00          & 0.00          & 0.00          & \textbf{1.00} & 0.00           & 0.00          & 0.00          & 0.00          & 0.00          & 0.00          & 0.00          & 0.00          & 0.00          \\
\hline

\end{tabular}}
\caption{The confusion matrix achieved by the EfficientNet-b0 model on the test subset of the CottonWeeds dataset.}
\label{tab:EfficientNet-b0}
\end{table*}

\begin{table*}[!th]
\centering
\renewcommand{\arraystretch}{1.5}
\resizebox{1.0\textwidth}{!}{
\begin{tabular}{c|c|c|c|c|c|c|c|c|c|c|c|c|c|c|c}
\hline
\multicolumn{1}{l|}{} & Carpetweeds   & Crabgrass     & Eclipta       & Goosegrass    & Morningglory  & Nutsedge      & PalmerAmaranth & PricklySida   & Purslane      & Ragweed       & Sicklepod     & SpottedSpurge & SpurredAnoda  & Swinecress    & Waterhemp     \\ \hline
Carpetweeds            & \textbf{1.00} & 0.00          & 0.00          & 0.00          & 0.00          & 0.00          & 0.00           & 0.00          & 0.00          & 0.00          & 0.00          & 0.00          & 0.00          & 0.00          & 0.00          \\ \hline
Crabgrass              & 0.00          & \textbf{0.94} & 0.00          & 0.06          & 0.00          & 0.00          & 0.00           & 0.00          & 0.00          & 0.00          & 0.00          & 0.00          & 0.00          & 0.00          & 0.00          \\ \hline
Eclipta                & 0.00          & 0.00          & \textbf{0.92} & 0.08          & 0.00          & 0.00          & 0.00           & 0.00          & 0.00          & 0.00          & 0.00          & 0.00          & 0.00          & 0.00          & 0.00          \\ \hline
Goosegrass             & 0.00          & 0.00          & 0.00          & \textbf{1.00} & 0.00          & 0.00          & 0.00           & 0.00          & 0.00          & 0.00          & 0.00          & 0.00          & 0.00          & 0.00          & 0.00          \\ \hline
Morningglory           & 0.00          & 0.00          & 0.00          & 0.00          & \textbf{0.99} & 0.00          & 0.01           & 0.00          & 0.01          & 0.00          & 0.00          & 0.00          & 0.00          & 0.00          & 0.00          \\ \hline
Nutsedge               & 0.00          & 0.00          & 0.00          & 0.00          & 0.00          & \textbf{1.00} & 0.00           & 0.00          & 0.00          & 0.00          & 0.00          & 0.00          & 0.00          & 0.00          & 0.00          \\
\hline

\end{tabular}}
\caption{The confusion matrix achieved by the EfficientNet-b1 model on the test subset of the CottonWeeds dataset.}
\label{tab:EfficientNet-b1}
\end{table*}

 \clearpage
\begin{table*}[!th]
\centering
\renewcommand{\arraystretch}{1.5}
\resizebox{1.0\textwidth}{!}{
\begin{tabular}{c|c|c|c|c|c|c|c|c|c|c|c|c|c|c|c}
\hline
\multicolumn{1}{l|}{} & Carpetweeds   & Crabgrass     & Eclipta       & Goosegrass    & Morningglory  & Nutsedge      & PalmerAmaranth & PricklySida   & Purslane      & Ragweed       & Sicklepod     & SpottedSpurge & SpurredAnoda  & Swinecress    & Waterhemp     \\ \hline
Carpetweeds            & \textbf{1.00} & 0.00          & 0.00          & 0.00          & 0.00          & 0.00          & 0.00           & 0.00          & 0.00          & 0.00          & 0.00          & 0.00          & 0.00          & 0.00          & 0.00          \\ \hline
Crabgrass              & 0.00          & \textbf{0.94} & 0.00          & 0.06          & 0.00          & 0.00          & 0.00           & 0.00          & 0.00          & 0.00          & 0.00          & 0.00          & 0.00          & 0.00          & 0.00          \\ \hline
Eclipta                & 0.00          & 0.00          & \textbf{0.92} & 0.08          & 0.00          & 0.00          & 0.00           & 0.00          & 0.00          & 0.00          & 0.00          & 0.00          & 0.00          & 0.00          & 0.00          \\ \hline
Goosegrass             & 0.00          & 0.00          & 0.00          & \textbf{1.00} & 0.00          & 0.00          & 0.00           & 0.00          & 0.00          & 0.00          & 0.00          & 0.00          & 0.00          & 0.00          & 0.00          \\ \hline
Morningglory           & 0.00          & 0.00          & 0.00          & 0.00          & \textbf{0.99} & 0.00          & 0.01           & 0.00          & 0.01          & 0.00          & 0.00          & 0.00          & 0.00          & 0.00          & 0.00          \\ \hline
Nutsedge               & 0.00          & 0.00          & 0.00          & 0.00          & 0.00          & \textbf{1.00} & 0.00           & 0.00          & 0.00          & 0.00          & 0.00          & 0.00          & 0.00          & 0.00          & 0.00          \\
\hline

\end{tabular}}
\caption{The confusion matrix achieved by the EfficientNet-b2 model on the test subset of the CottonWeeds dataset.}
\label{tab:EfficientNet-b2}
\end{table*}

\begin{table*}[!th]
\centering
\renewcommand{\arraystretch}{1.5}
\resizebox{1.0\textwidth}{!}{
%
}
\caption{The confusion matrix achieved by the EfficientNet-b3 model on the test subset of the CottonWeeds dataset.}
\label{tab:EfficientNet-b3}
\end{table*}

\begin{table*}[!th]
\centering
\renewcommand{\arraystretch}{1.5}
\resizebox{1.0\textwidth}{!}{
%
}
\caption{The confusion matrix achieved by the EfficientNet-b5 model on the test subset of the CottonWeeds dataset.}
\label{tab:EfficientNet-b5}
\end{table*}

\section{Confusion Matrices of Deep learning models (WCE Loss)}

\bibliography{ref}{}
\bibliographystyle{plain}